\documentclass{article} 
\usepackage{authblk}
\usepackage[T1]{fontenc}
\usepackage{geometry}
\usepackage{float}
\usepackage{amsmath}
\usepackage{nameref}
\usepackage{algorithm}
\usepackage{amsthm}
\usepackage{amssymb}
\usepackage{comment}

\usepackage{graphicx}
\usepackage{bm}

\theoremstyle{example}
\newtheorem*{example}{Example}

\theoremstyle{definition}
\newtheorem*{definition}{Definition}
\newcommand{\x}{\bm{x}}

\newcommand{\bb}[1]{\mathbb{#1}}
\newcommand{\cov}{\text{cov}}

\title{Imputing Missing Observations with Time Sliced Synthetic Minority Oversampling Technique}

\author[1]{Andrew Baumgartner,  Sevda Molani, Qi Wei \& Jennifer Hadlock \\
\small{Institute for Systems Biology, Seattle, WA}}
\begin{document}

\maketitle
\abstract{We present a simple yet novel time series imputation technique with the goal of constructing an irregular time series that is uniform across every sample in a data set. Specifically, we fix a grid defined by the midpoints of non-overlapping bins (dubbed "slices") of observation times and ensure that each sample has values for all of the features at that given time. This allows one to both impute fully missing observations to allow uniform time series classification across the entire data and, in special cases, to impute individually missing features. To do so, we slightly generalize the well-known class imbalance algorithm SMOTE \cite{smote} to allow component wise nearest neighbor interpolation that preserves correlations when there are no missing features.  We visualize the method in the simplified setting of 2-dimensional uncoupled harmonic oscillators. Next, we use tSMOTE to train an Encoder/Decoder long-short term memory (LSTM) model with Logistic Regression for predicting and classifying distinct trajectories of different 2D oscillators. After illustrating the the utility of tSMOTE in this context, we use the same architecture to train a clinical model for COVID-19 disease severity on an imputed data set.  Our experiments show an improvement over standard mean and median imputation techniques by allowing a wider class of patient trajectories to be recognized by the model, as well as improvement over aggregated classification models. } 

%\keywords{time series, imputation, SMOTE, class imbalance, LSTM, COVID-19}

%\jnlcitation{\cname{%
%\author{Andrew Baumgartner} and
%\author{Jennifer Hadlock}} (\cyear{<year>}),
%\ctitle{<journal title>}, \cjournal{<journal name>} <year> <vol> Page <xxx>-<xxx>}
%\footnotetext{\textbf{<abbreviation head:>} <abbreviations> ..} 

\section*{Introduction}\label{intro}

In this work we address common issues which are present in problems involving
time series data sets: 1) individual samples can have a different number
of observations, 2) the time between consecutive observations is not uniform across
the data set, 3) the time between consecutive observations is not uniform across
each sample and 4) each observation may not contain values for every feature. See Fig. \ref{fig:dataviz} for a visual schematic.  These issues make it difficult to infer the trajectory of samples
in feature space, an important component for being able to predict the outcome of
a given classification scheme to an arbitrary (but bounded) time in the future.
To address these issues, we developed a time series imputation technique known
as time sliced minority oversampling technique, or tSMOTE. The idea is to partition a time interval (depending on the problem at hand) into a fixed number of subintervals,
dubbed ``time slices". These time slices are of varying length and contain an approximately equal number
of observations from at least two different samples. We then perform SMOTE \cite{smote} on each feature
in a time slice and impute these new samples into the trajectories
of those that are missing it.  Additionally, when the variables are uncorrelated,  we can impute the values of individually missing null features at a given time point. The details of our technique are presented in Results and Methods.

Treating irregularly sampled time series as a missing data problem is not a new idea. To date there have been many approaches to solving some subset of the issues stated above.  A  majority of them are similar, but different in aim. Most existing techniques impute missing features of existing observations. This can be done both through classical imputation techniques like  k-nearest neighbor (KNN) imputation \cite{knn}, matrix approaches such as singular value decomposition (SVD) \cite{mazumder2010spectral}, principle component analysis (PCA) \cite{PCA}  or matrix factorization \cite{koren2009matrix}, and chain equations \cite{azur2011multiple}. When applied to time series, these approaches do not fully leverage time dependent information and may not be applicable given the distribution of observations. To address these issues, more recent attempts have leveraged neural network architectures which were developed for sequence-to-sequence learning tasks. These include recurrent neural networks (RNN) \cite{alzModel, lipton2016directly, lipton2016modeling, che2018recurrent, kim2018temporal} and their generalizations such as bi- and multi-directional RNNs \cite{cao2018brits, yoon2017multi}, Long Short-Term Memory (LSTMs) \cite{yuan2018imputation, AAAIW1817154}, attention networks \cite{singh2019multi, nguyen2017deep}, generative adversarial networks (GAN) \cite{multiGAN} and gated recurrent units (GRU) \cite{che2018recurrent}, sometimes with the addition of global interpolation layers \cite{shukla2019interpolation}. These approaches use the time dependent structure and frequency of the missing data as features in the models to help facilitate the learning process. These approaches also have the advantage of being able to impute the data and perform classification or regression tasks within the same architecture which is useful for streamlining analysis.  However, extracting the fully imputed data set is not straightforward, making their use in studying individual trajectories limited.  Other approaches to the problem are more probabilistic in nature, treating the time series as a Gaussian process and employing kernel techniques  \cite{futoma2017learning, li2015classification, marlin2012unsupervised}. This particular set, while powerful, does not give you a fully imputed time series as an output but instead performs the required task directly on the sparse data set.

Below we present tSMOTE and illustrate its use in a variety of settings. First, we look at a toy model and use tSMOTE to impute missing observations from a 2D simple harmonic oscillator.  We illustrate here explicitly how the time slices show up, how they change size depending on the distribution of samples, and examine the similarity between the synthetic data and the original data. Next we use tSMOTE to build and train two models for classification tasks: one model to distinguish trajectories of two distinct 2D harmonic oscillators, where we compare our technique to imputing the mean and median of each time slice, and one  on electronic health records (EHRs) of patients with COVID-19 to build a severity prediction model, where we compare to models in which time series information has been aggregated.  Once the full trajectory of each sample has been imputed,  we train an Encoder-Decoder LSTM model \cite{sutskever2014sequence, cho2014learning} to predict trajectories of new samples and classify their endpoints using logistic regression. We conclude with a discussion as well as ideas for future work. 

\begin{comment}
Finally,  we use tSMOTE in a completely novel setting wherein direct verification remains difficult: computing proteomic velocities for COVID-19 patients.  We use tSMOTE to generate a full proteomic expression profile for each patient over a 50-day period, and then directly compute the derivatives for each patients using a combination of smoothing kernels, and simple forward difference equations.  Here we only wish to illustrate a proof-of-concept and hope to return later for more detailed analysis and verification of the findings. 
\end{comment}

\section*{Results}\label{sec:results}
As stated in the introduction our
algorithm leverages the well known SMOTE algorithm \cite{smote} developed for class imbalance
problems to generate new samples in each time slice.  The idea is to use linear interpolation between
a sample and its $n$ nearest neighbors to generate synthetic data
points. Linear interpolation may not be appropriate for samples which lie on a low dimensional,
non-Euclidean subspace of $\mathbb{R}^{n_{F}}$, but if the distance from $\bm{x}_{i}$
to its neighbor $\bm{x}_{j}$ is small compared to the scale set by the intrinsic
curvature of the data manifold, the approximation can be reliable. 

\begin{figure}[t]
\centering
\includegraphics[scale=0.5]{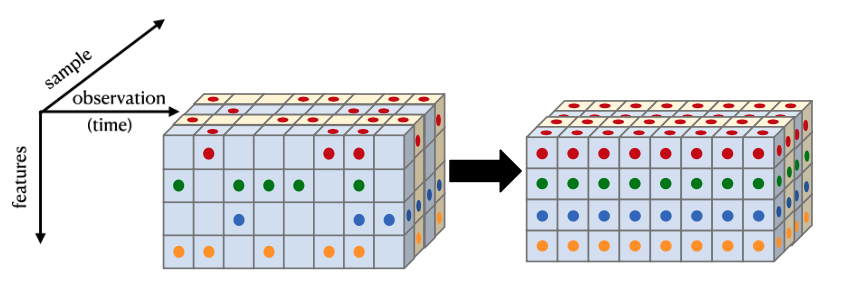}
\caption{Rough visualization of our data set before and after tSMOTE has been applied. Colored dots represent non-null entries, which can occur across all features at a given time, or some subset of the features.  Each color represents a different feature in the data set.  The goal is to ensure each sample has the same features observed at the same times.}\label{fig:dataviz}
\end{figure}

One important thing to remember is that our data is no longer organized as a typical 2-dimensional matrix of samples and features, but instead becomes a three-dimensional tensor of samples, features and time, as illustrated in Fig. \ref{fig:dataviz}.  Throughout, we use the term ``observation" when referring the individual time points within a given sample.  The goal of tSMOTE is to give each sample the same number of observations at the same times.  
 
 \begin{comment} 
 In table \ref{tab:example} we present the samples, features and observations specific to the examples presented in this work.
 \end{commen\t}

\begin{comment}
Before we proceed we also note that tSMOTE does not have an optimal procedure for imputing the missing observations. As such, the resulting trajectories are not (yet) meant to mimic the true trajectories of the samples.  The best use case for tSMOTE is generating observations which can be used to train a relevant classifier, such as an RNN, for increased accuracy. This is similar in spirit to noise injection techniques \cite{NNnoise} where new training samples are created from adding noise to existing samples. Alternatively, tSMOTE can also help increase the sample sizes in a given time slice to increase the power of statistical tests done in that time slice without significantly altering the first and second moments of the data (see Methods for details).  We hope to improve upon the interpretability of individual trajectories in forthcoming work. 
\end{comment}

\begin{comment}
\begin{table}[t]
\begin{centering}
\begin{tabular}{|c|c|c|c|}
\hline
& \textbf{2D Oscillator} & \textbf{COVID prediction} & \textbf{Proteomic Velocity}
 \tabularnewline
\hline
\hline 
\textbf{Sample} & generated samples & patient & patient
\tabularnewline
\hline
\textbf{Features} & x-and y- coordinates & vital signs \& laboratory results  & protein expression \tabularnewline
\hline
\textbf{Observations} & sampled from various distribution & measurements by clinical staff &  blood draws \tabularnewline
\hline
\end{tabular}
\caption{Specific samples, features and observations for the examples presented in this work. }
\label{tab:example}
\par\end{centering}
\end{table}
\end{comment}

tSMOTE is concerned with data sets in which each sample is a time series of the form

\[
\bm{x}_{i}(t)=\left(\bm{x}_{i}(t_{1}^{i}),\bm{x}_{i}(t_{2}^{i}),...,\bm{x}_{i}(t_{m_{i}}^{i})\right)
\]
where $\bm{x}_{i}(t_{\mu}^{i})$ denotes the feature vector of sample $i$ at time
$t_{\mu}^{i}$, with component $\left[\bm{x}_{i}(t_{\mu}^{i})\right]_{k}$. Here, the $i$ superscript ties
the observation to the corresponding sample , while the subscript $\mu$ denotes
where on the trajectory it lies. Let the size of the data set be $n_{D}$ and the
number of time slices be $n_{T}$. We denote the number of observations associated
to $\bm{x}_{i}$ as $m_{i}$, and the dimension of $\bm{x}_{i}$ (equivalently the number
of features) as $n_{F}$. With this notation, the issues mentioned in the introduction
can be formalized in the following way: 
\begin{enumerate}
\item Each sample may contain a different number of observations: $m_{i}\ne m_{j}$ for some
$i\ne j$. 
\item The time elapsed between consecutive observations is non-uniform across the data set:
$|t_{\mu+1}^{i}-t_{\mu}^{i}|\ne|t_{\mu+1}^{j}-t_{\mu}^{j}|$ for some $i\ne j$
\item The time elapsed between consecutive observations is non-uniform across each data
point: $|t_{\mu+1}^{i}-t_{\mu}^{i}|\ne|t_{\nu+1}^{i}-t_{\nu}^{i}|$ for some $\mu\ne\nu$
\item $\left[\bm{x}_{i}(t_{\mu}^{i})\right]_{k}=\text{Null}$ for some $i$.
\end{enumerate}
tSMOTE was designed to solve issues 1) and 2) with this approach, leaving issues 3) as a feature, not an issue.  The irregularity of the time intervals is of secondary concern, since one can always compute what the time interval is from the initial slicing procedure and supplement any analysis with this information. As for issue 4), tSMOTE can be applied reliably if the features are known to be independent. The addition of nulls will destroy the correlations between variables, since the nearest neighbors along each feature direction will generally be different.  

\subsection*{tSMOTE}\label{sec:tSMOTE}

The tSMOTE algorithm works in three steps: 
\begin{enumerate}
\item Partition the time interval such that every time slice has an (approximately) equal number of points.
\item Generate synthetic samples within each time slice using SMOTE  \cite{smote} along each feature direction individually.
\item Sample (with or without replacement) from synthetic samples and impute synthetic observations into samples which do not have observations at that time.
\end{enumerate}
The result of this procedure is a data set in which every sample has each observation at each time. There is an optional fourth step wherein one applies a smoothing filter, such as a nonuniform Savitzky-Golay filter \cite{savgol}, to each sample to further reduce the noise and isolate the signal.  This may not be preferable depending on the problem at hand, but is easily accommodated in our code. 

There are three important things to keep in mind when preparing a data set for classification using tSMOTE: 1) the
time slices themselves must be constructed on the entire data set, so that each class has their time series defined at the same time points, 2) tSMOTE must be performed on each class separately so that the resulting imputed time series behaves like others in the class and 3) when sampling without replacement, one must generate enough synthetic samples to impute all the data, i.e. one needs greater than $num \, samples - num. \, tot. \,obs/num\,slices$ samples in each slice. 

This approach is not without it's limitations. First and foremost the whole procedure relies on assuming the dynamics of the system is ergodic, so that we may determine the time-dependent behavior of the system via ensemble
averages. This may not be the case in many real world data sets, but is a reasonable
approximation for systems which can be described by deterministic dynamical systems.

Additionally,  interpolating along each axis individually can destroy correlations between variables when null values are present. To see this, consider this procedure as a non-parametric sampling technique within each time slice. By interpolating along each direction individually, you are picking each component from it's marginal distribution. This will wash out any correlation among the variables. However, when there are no nulls present, the resulting distribution can still preserve correlations since the nearest neighbors along each feature direction will likely be contained in the same small neighborhood surrounding the initial point from which they were created.  By reassembling these samples from the marginal distributions in such a way that each individual variable is drawn from the same neighborhood, correlations will still be preserved in the final synthetic set. On the other hand, when nulls are present, one is required sample from beyond these nearest neighborhoods in each direction, since the would-be nearest neighbors may not have a component in that slice.  As such, the resulting vectors are reassembled using components from neighborhoods of different data points, thus destroying the correlations. If your variables are independent, then this is a perfectly valid procedure and tSMOTE can be applied to data with nulls without worry.

\subsection*{Experiments}
We now turn to some experiments which demonstrate the utility to tSMOTE.  First, we start with toy examples of uncoupled 2D oscillators. This example is meant to help visualize the synthetic data generated by tSMOTE and how the sampling distribution affects the size and location of these slices. Next we use an Encoder-Decoder LSTM \cite{sutskever2014sequence, cho2014learning} coupled with logistic regression in the 2D oscillator setting to discriminate between oscillators with a different ratio of frequencies. We show a large improvement with tSMOTE over naive mean and median imputation of each slice, as well as over an aggregate classification model. We then apply this model to COVID-19 severity prediction, where we again show improved accuracy and area under the receiver operating characteristic curve (AUC) an aggregated model. 

\subsubsection*{Reconstructing Trajectories of 2D Harmonic Oscillators \label{subsec:tSMOTE-in-Action}}

\begin{figure}[h]
\begin{minipage}[t]{0.3\columnwidth}%
\includegraphics[scale=0.19]{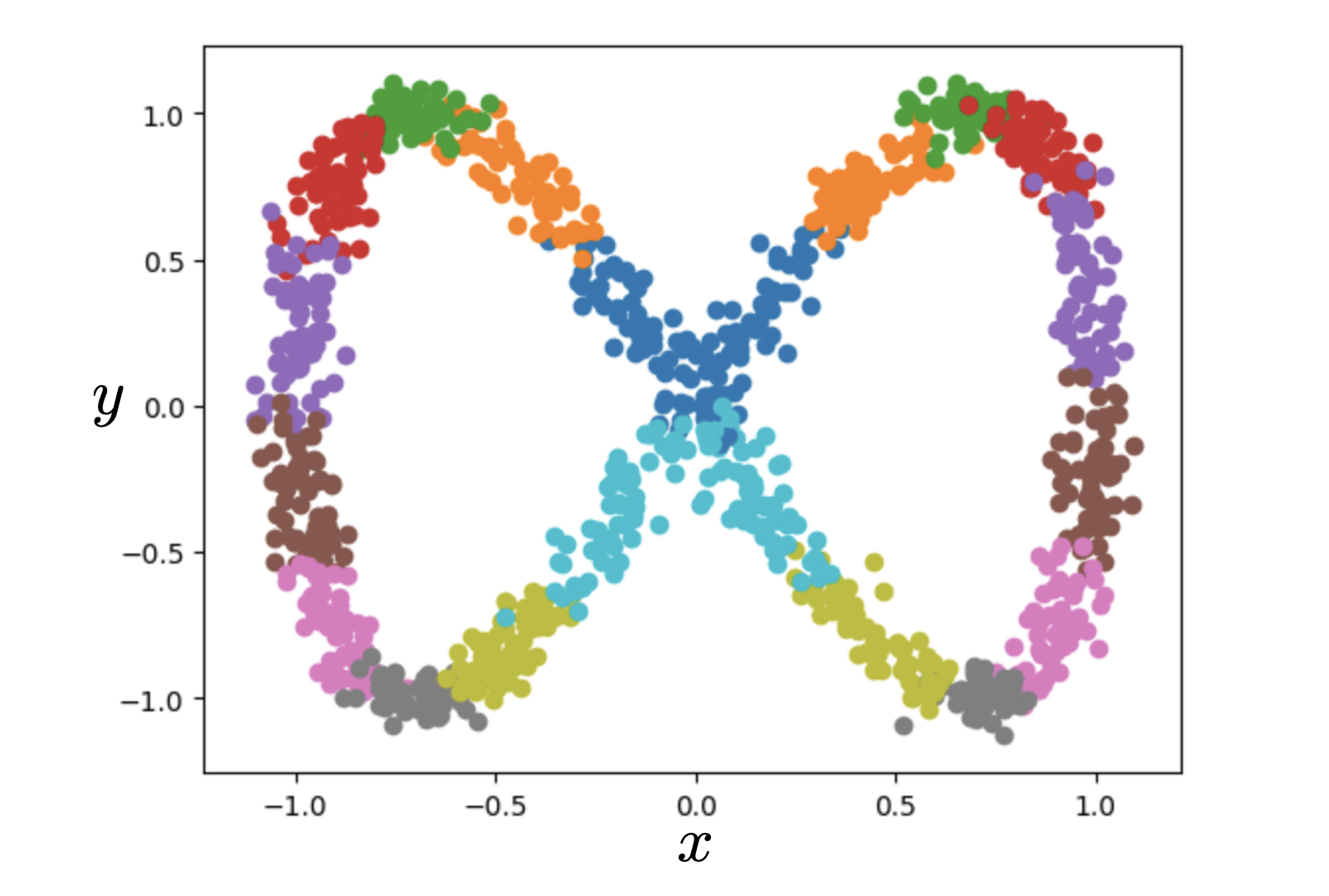}
\begin{center}
(a)
\par\end{center}%
\end{minipage}%
\begin{minipage}[t]{0.3\columnwidth}%
\includegraphics[scale=0.25]{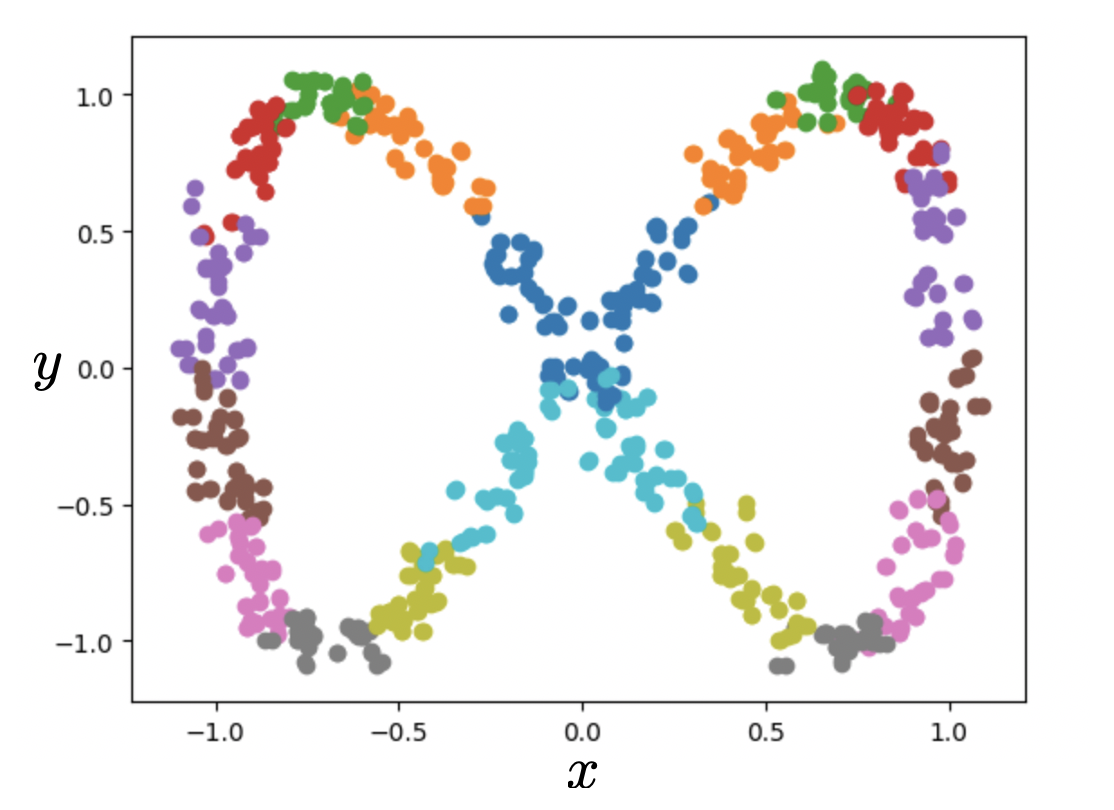}
\begin{center}
(b)
\par\end{center}%
\end{minipage}
\centering
\begin{minipage}[t]{0.3\columnwidth}%
\includegraphics[scale=0.25]{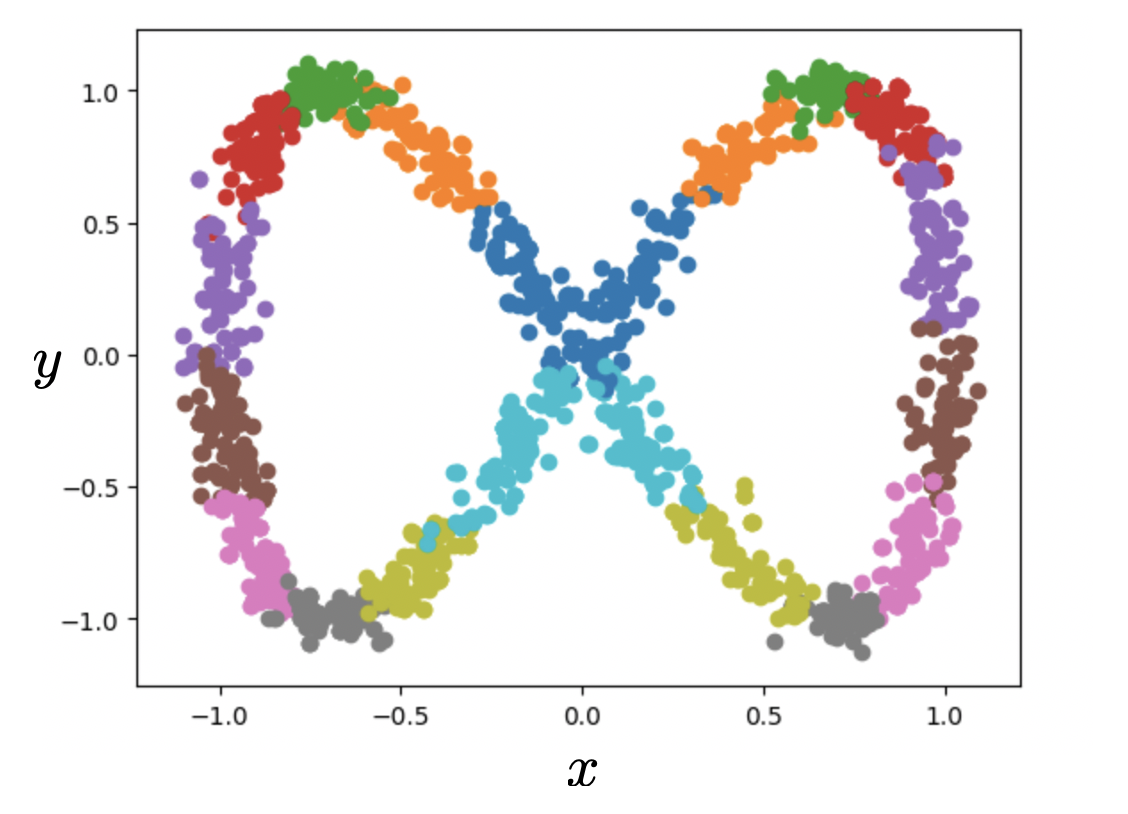}
\begin{center}
(c)
\par\end{center}%
\end{minipage}

\begin{minipage}[t]{0.3\columnwidth}%
\includegraphics[scale=0.19]{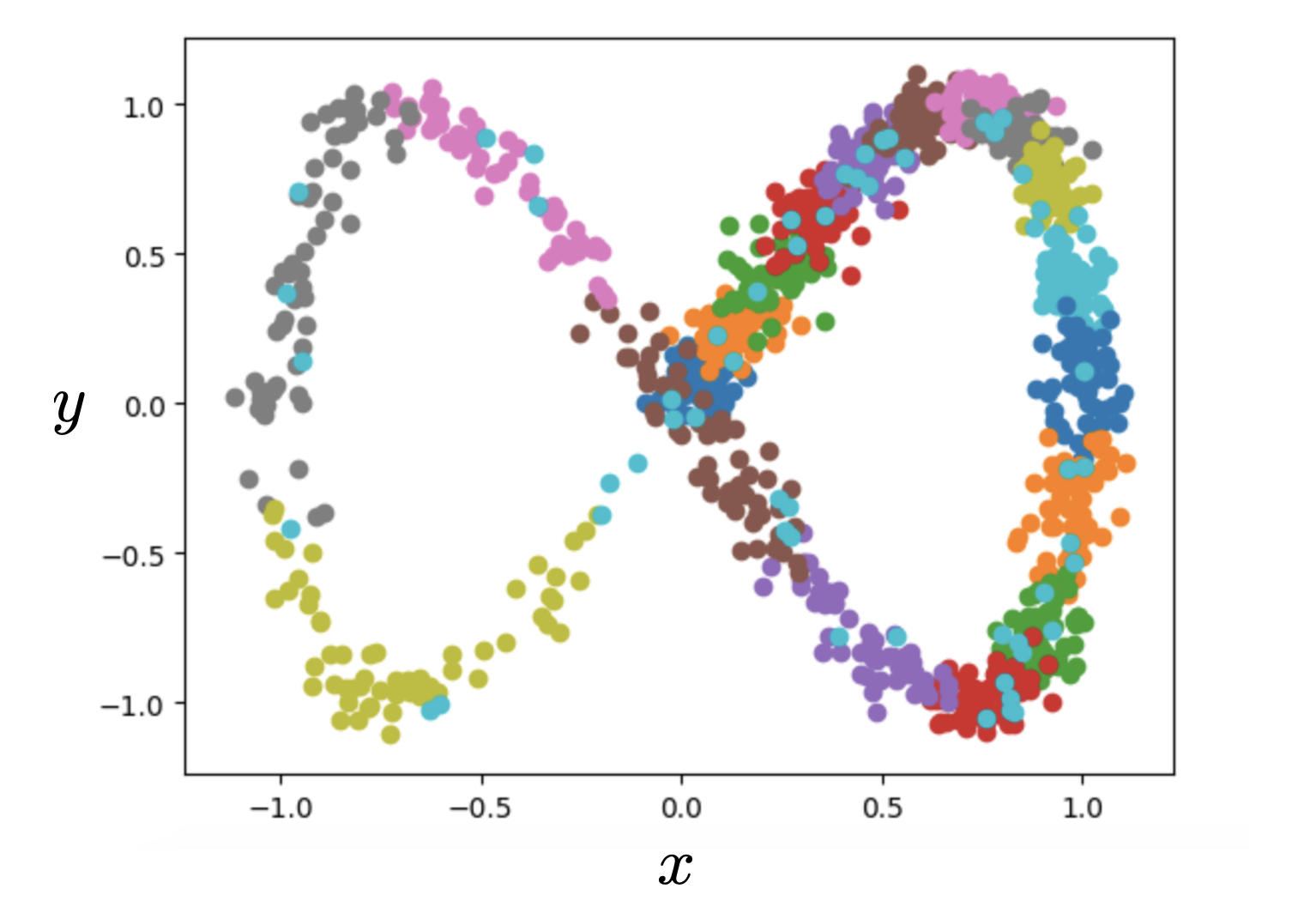}
\begin{center}
(d)
\par\end{center}%
\end{minipage}%
\begin{minipage}[t]{0.3 \columnwidth}%
\includegraphics[scale=0.25]{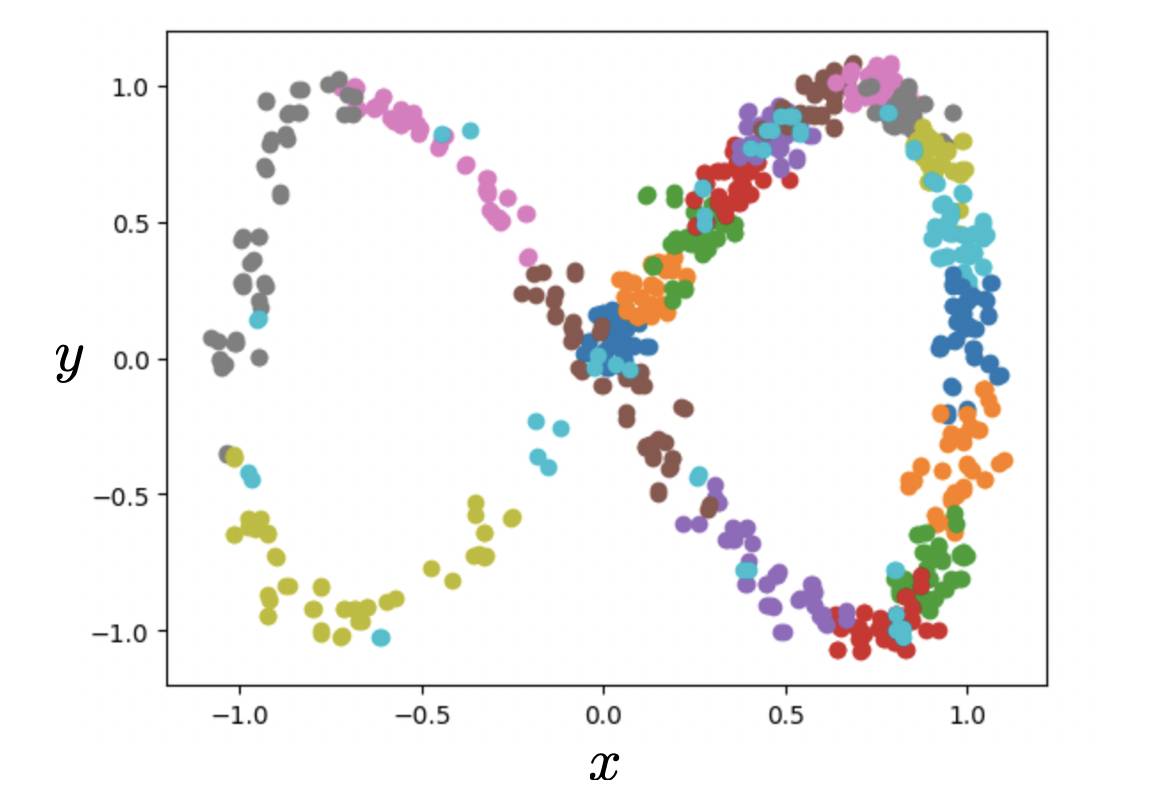}
\begin{center}
 (e)
\par\end{center}%
\end{minipage}
\centering
\begin{minipage}[t]{0.30\columnwidth}%
\includegraphics[scale=0.25]{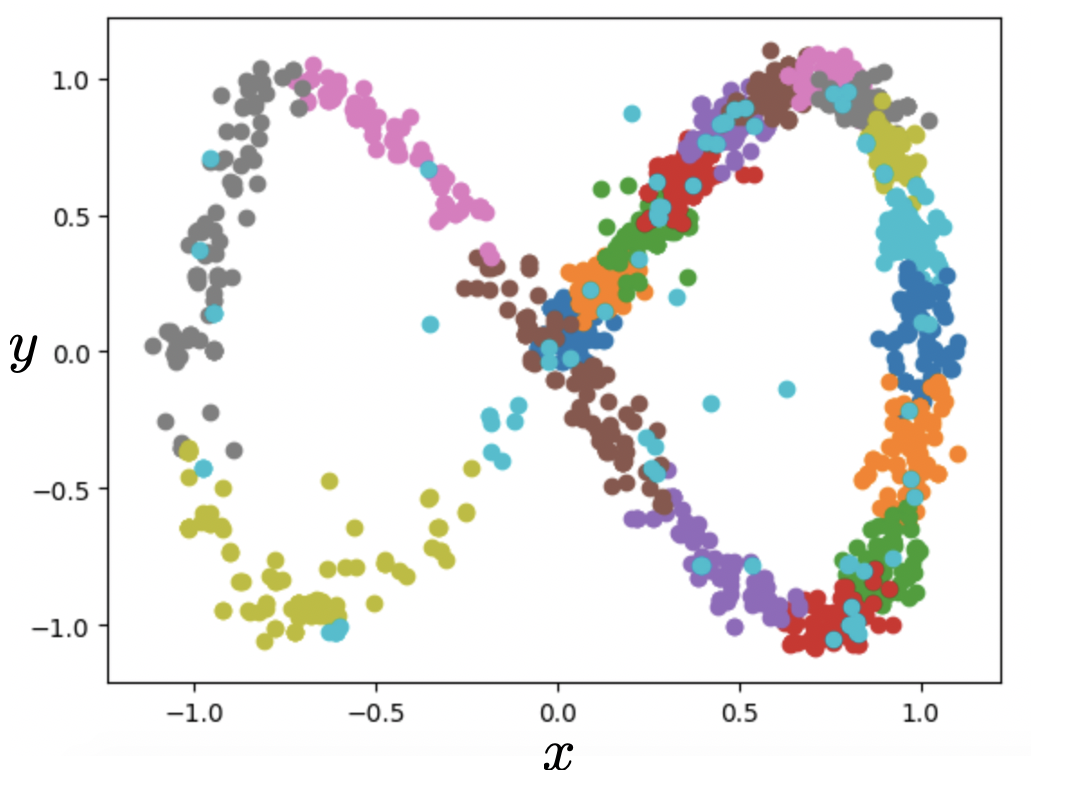}
\begin{center}
(f)
\par\end{center}%
\end{minipage}\hfill{}%

\caption{tSMOTE illustration for a 2D harmonic oscillator with observations sampled from a
uniform distribution (a)-(c) and an exponential distribution (d)-(f). (a) \& (d) the original data set generated as described in the text (b) \& (e) synthetic data generated by tSMOTE and (c) \& (f) the
original data set after imputation.}
\label{fig:uniformOsc} 
\end{figure}

As a toy example let us see how well tSMOTE can impute in a simple case: a 2D harmonic oscillator.
This is a system defined by 
\begin{equation}\label{eq:LC}
\begin{split}
x & =\sin(\omega_{x}t)+\epsilon_{x}\\
y & =\sin(\omega_{y}t+\delta)+\epsilon_{y}\\
t_{min} & =0\\
t_{max} & =\max\left\{ \frac{2\pi}{\omega_{x}},\frac{2\pi}{\omega_{y}}\right\} 
\end{split}
\end{equation}
where $\epsilon$ represents noise drawn from a Gaussian distribution with $\mu=0$ and
$\sigma=.05$ . For simplicity, we take $\delta=0$ and $\omega_{x}/\omega_{y}=1/2$.
To simulate the irregularity of observations present in real world data, we randomly
sample a number of times from various distributions and plug them into $\bm{x}(t)$.
Our resulting data set has 100 samples, with anywhere from 5 to 20 observations per
sample at irregularly spaced intervals. Our goal is to see how well our interpolation and nearest neighbor imputation
preserves the structure of the original data, and how time slice size  affects the synthetic data. The original and synthetic data are shown in Fig.  \ref{fig:uniformOsc} for observations drawn from a uniform distribution and exponential distribution,respectively. In both figures, we display both the original and synthetically generated data, as well as the data set after imputation. The observations are calculated to be the midpoints of the appropriate time slice. As can be seen, the synthetically generated data very much resembles the initial data set for both the uniform and exponentially distributed observations.  

\subsubsection*{Differentiating 2D oscillators}

\begin{figure}[h]
\begin{centering}
\includegraphics[scale=0.5]{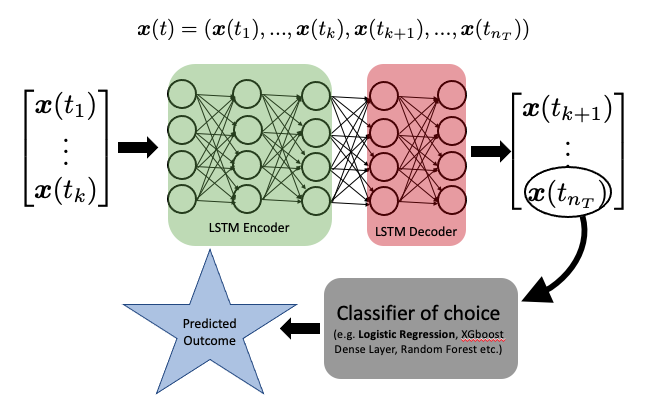}
\par\end{centering}
\caption{Schematic of our prediction model. We input the first $k$ times points,  and predict the next $n_T-k$ time points. We then use a classifier (logistic regression for this work) to classify the end point. }
 \label{fig:predict}
\end{figure}

\begin{figure}[t]
\begin{minipage}[t]{0.45\columnwidth}%
\begin{center}
\includegraphics[scale=0.6]{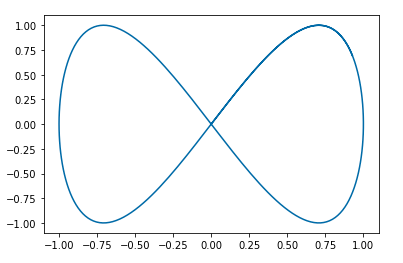}
\par\end{center}
\begin{center}
(a)
\par\end{center}%
\end{minipage}\hfill{}%
\begin{minipage}[t]{0.45\columnwidth}%
\begin{center}
\includegraphics[scale=0.6]{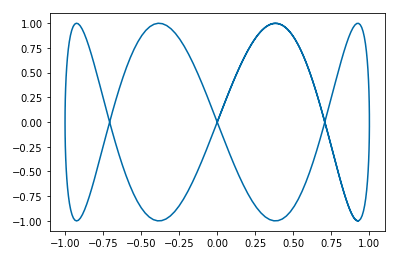}
\par\end{center}
\begin{center}
(b)
\par\end{center}%
\end{minipage}

%\begin{table}[t]

\begin{centering}
\begin{tabular}{|c|c|c|c|}
\hline 
 & LSTM MSE & Logistic Regression AUC & Logistic Regression Accuracy\tabularnewline
\hline 
\hline 
tSMOTE & \textbf{1.43356} & \textbf{0.93358}  & \textbf{0.92780} \tabularnewline
\hline 
Mean & 4.76737 $\times 10^{7}$ &0.492250 & 0.48700\tabularnewline
\hline 
Median & 4.52381 $\times 10^{16}$  & 0.49958 & 0.48080 \tabularnewline
\hline 
\end{tabular}
\par\end{centering}
\caption{The 2D oscillator curves used for testing tSMOTE against mean and median imputation using the model shown in Fig. \ref{fig:predict}, and the resulting metrics compared to mean and median imputation.}\label{fig:curves}
%\end{table}
\end{figure}

We compare our tSMOTE imputation to two obvious alternatives--imputing with the mean and median of each time slice.  To make such a comparison, we use an Encoder-Decoder LSTM \cite{sutskever2014sequence, cho2014learning} to predict the latter part of the trajectory from the initial portion, and use logistic regression to differentiate two curves using the endpoint. These curves are defined by eq. \eqref{eq:LC} $\omega_y=2\omega_x$ and $\omega_y=4\omega_x$ (see Fig. \ref{fig:curves}). We generated our samples in the same was as the previous example, except with $\sigma=.1$, 270 and 180 samples on each curve respectively, and between 5 and 20 time points sampled from both the uniform and exponential distribution.  We used tSMOTE, mean and median imputation to bring each of these trajectories onto an irregular grid with 50 times points, where the median time of each slice is used as the grid point. After this, we generated 30 and 20 samples on the entire grid to use as the test set, so that we may compare each imputation method in a controlled way.  The structure of the model itself is shown in Fig. \ref{fig:predict}. The error metrics averaged over 100 runs are displayed in Fig. \ref{fig:curves}.  Additionally, we z-normalize the data prior to training, using the mean and standard deviation of the training set when normalizing the test set.  We note that when using the exponential distribution for sampling our time points, the tSMOTE imputation error metrics are comparable to the other two approaches. This is because the last time slice becomes parametrically larger than the other time slices due to the tails in the exponential distribution. This effect can be seen in Fig.  \ref{fig:uniformOsc}, where the light blue points from the final time slice span the entirety of the 2D oscillator curve.  As such, it makes it difficult for the classifier to accurately discriminate between the curves since last observation on each curve occurs in the time interval $[\sum_{i=1} ^{49} \delta t_i, \infty)$, thus placing it anywhere along the curve.

\subsubsection*{Application to COVID Severity Prediction}\label{sec:covPred}
\begin{comment}
\begin{figure}[t]
\begin{centering}
\includegraphics[scale=0.5]{Screen Shot 2021-05-24 at 3.32.11 PM.png}
\par\end{centering}
\caption{Distribution of observations as a function of the time slices}
\label{timeDist}
\end{figure}
\end{comment}

To further illustrate the power of tSMOTE, we use our procedure to impute missing
observations to improve prediction of COVID-19 disease severity. In particular,
we take as input a time series from the first 12 hours of a patient's hospital stay and attempt to predict the severity outcome over the next 7 days.   Our definition of severity is based on the WHO Ordinal Score (WOS) \cite{WOS},  which assigned a number from 1 (healthy) to 8 (dead) based primarily on the need for respiratory assistance. For our model, we stratify our cohort into ``moderate''
(WOS $<5$) and ``severe'' (WOS $\ge5$), where moderate patients are hospitalized and require either no breathing assistance, or a nasal canula, while severe patients have either a high flow nasal canula, or require mechanical ventilation, extracorporeal membrane oxygenation (ECMO),  vasopressor or some combination thereof.  To make our prediction, we restrict to patients with moderate COVID-19 in the first 12 hours of their hospital stay, and predict their maximum WHO score over the next week.

\begin{figure}[H]
\begin{minipage}[t]{0.23\columnwidth}%
\includegraphics[scale=0.24]{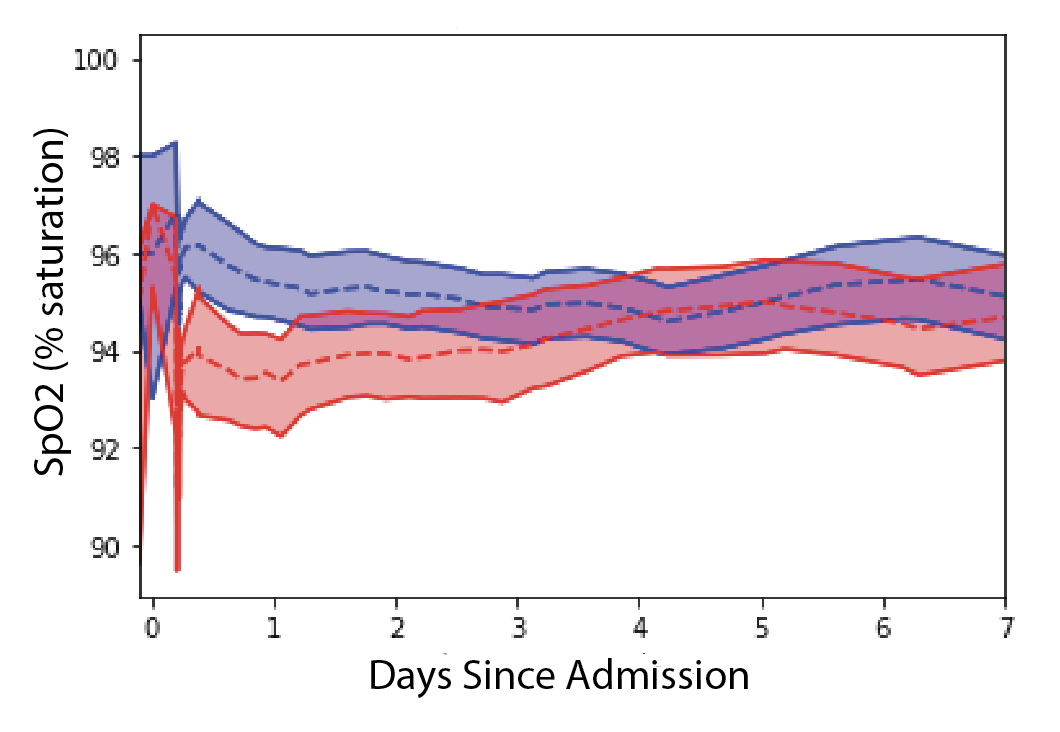}
\begin{center}
 
\par\end{center}%
\end{minipage}\hfill{}%
\begin{minipage}[t]{0.23\columnwidth}%
\includegraphics[scale=0.25]{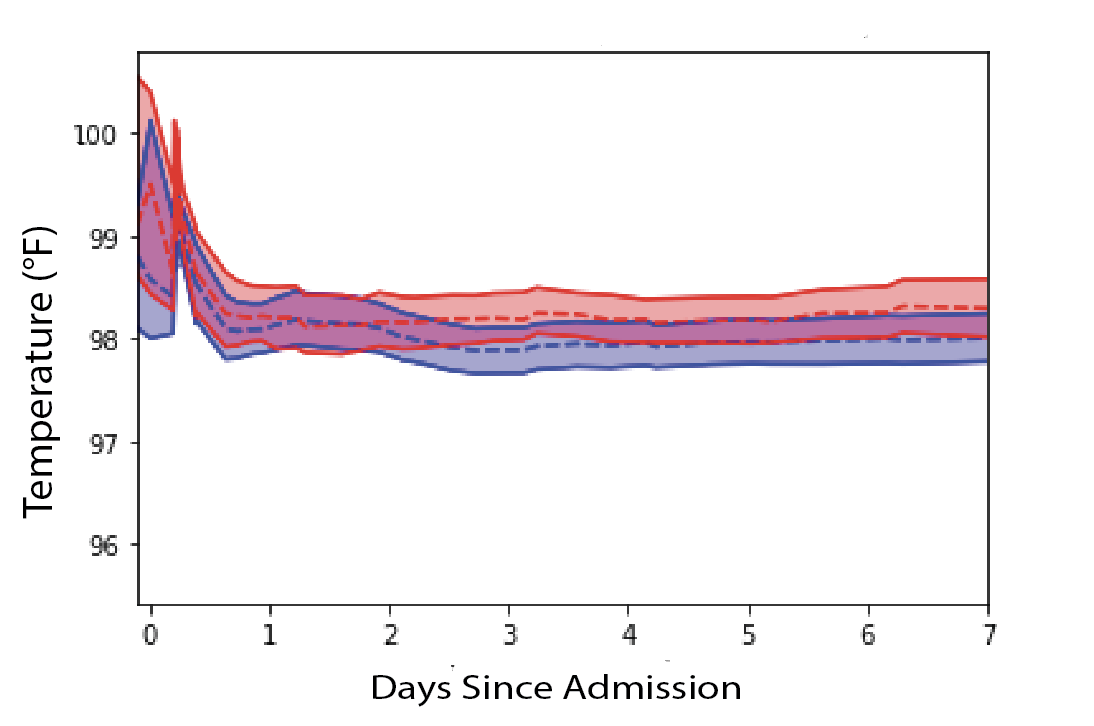}
\begin{center}
 
\par\end{center}%
\end{minipage}\hfill{}%
\begin{minipage}[t]{0.23\columnwidth}%
\includegraphics[scale=0.24]{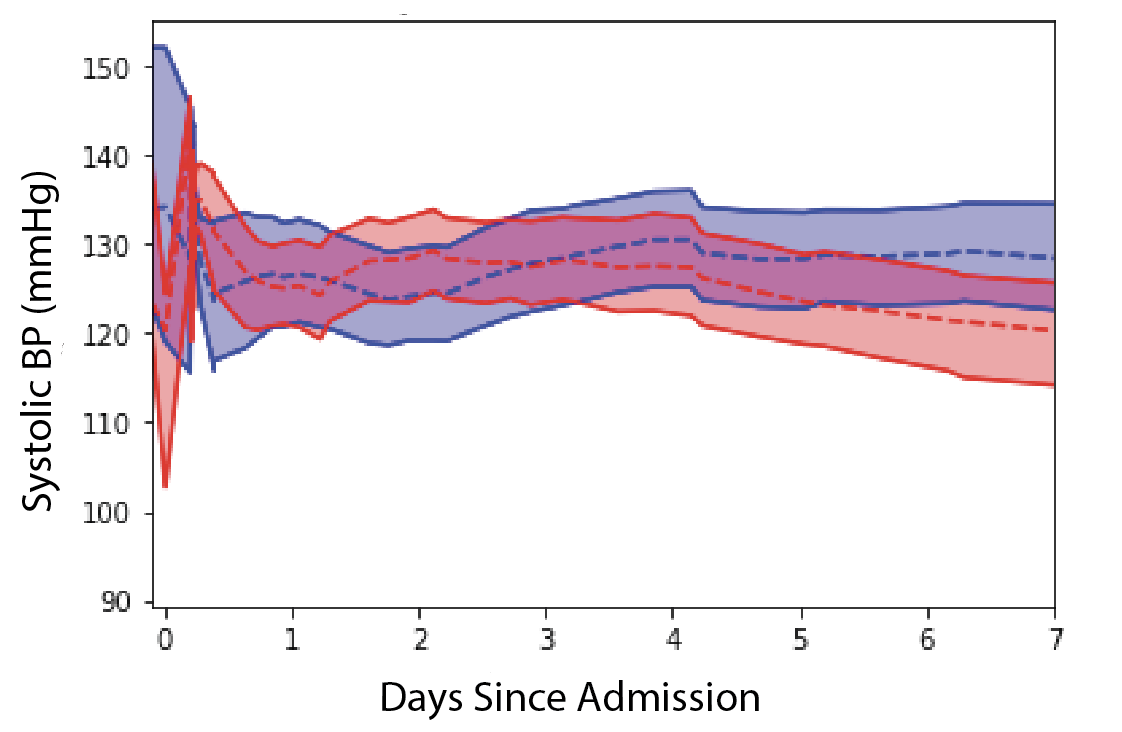}
\begin{center}
 
\par\end{center}%
\end{minipage}\hfill{}% 
\begin{minipage}[t]{0.23\columnwidth}%
\includegraphics[scale=0.24]{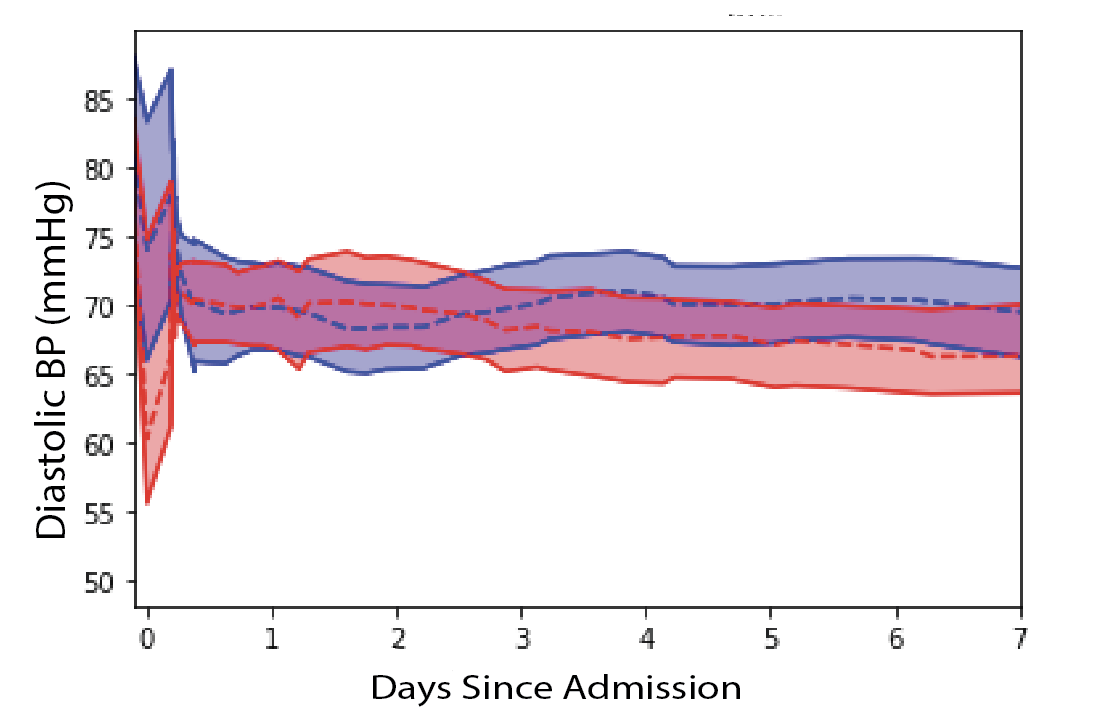}
\begin{center}
 
\par\end{center}%
\end{minipage}\hfill{}%

\begin{minipage}[t]{0.23\columnwidth}%
\includegraphics[scale=0.24]{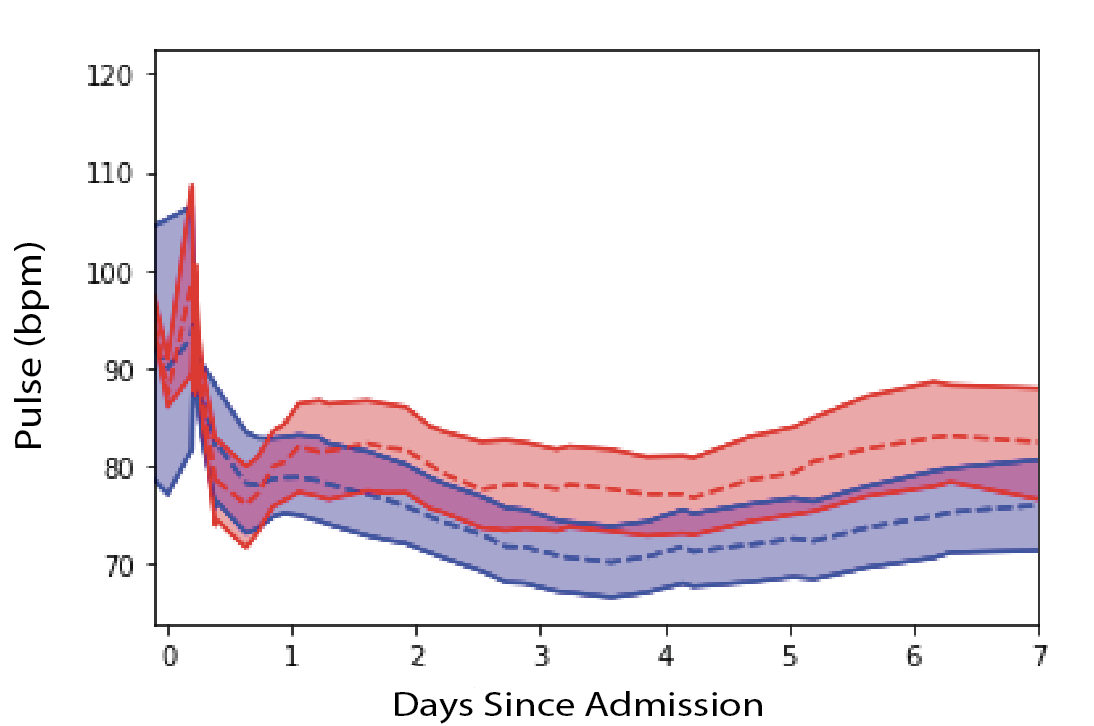}
\begin{center}
 
\par\end{center}%
\end{minipage}\hfill{}%
\begin{minipage}[t]{0.23\columnwidth}%
\includegraphics[scale=0.25]{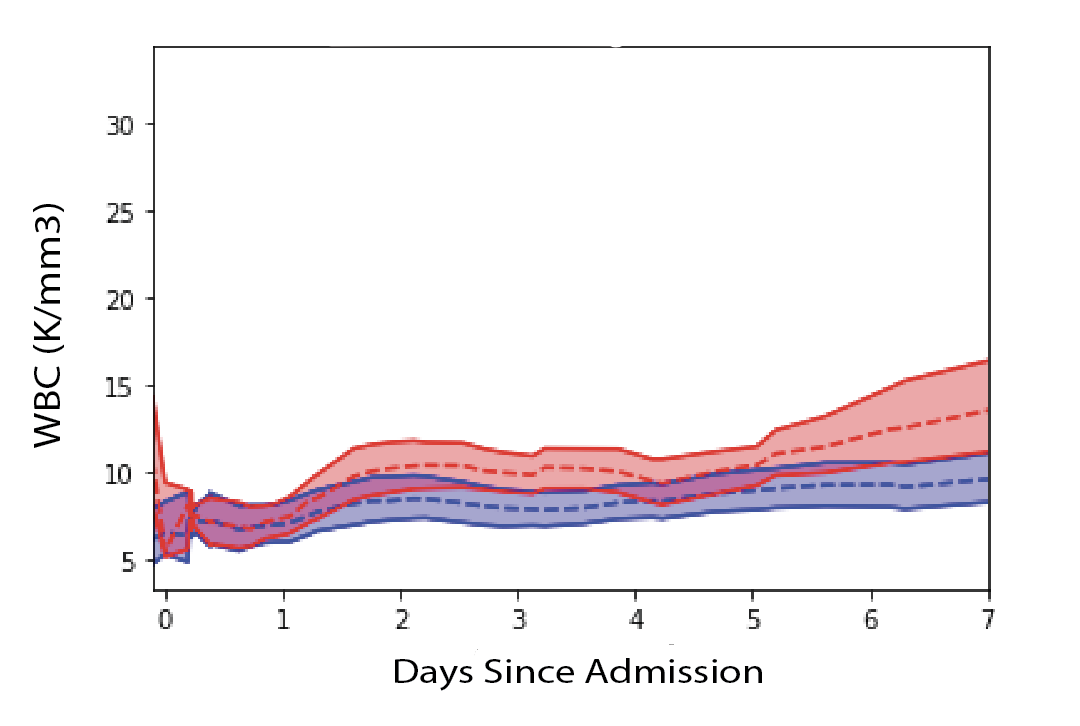}
\begin{center}
 
\par\end{center}%
\end{minipage}\hfill{}%
\begin{minipage}[t]{0.23\columnwidth}%
\includegraphics[scale=0.25]{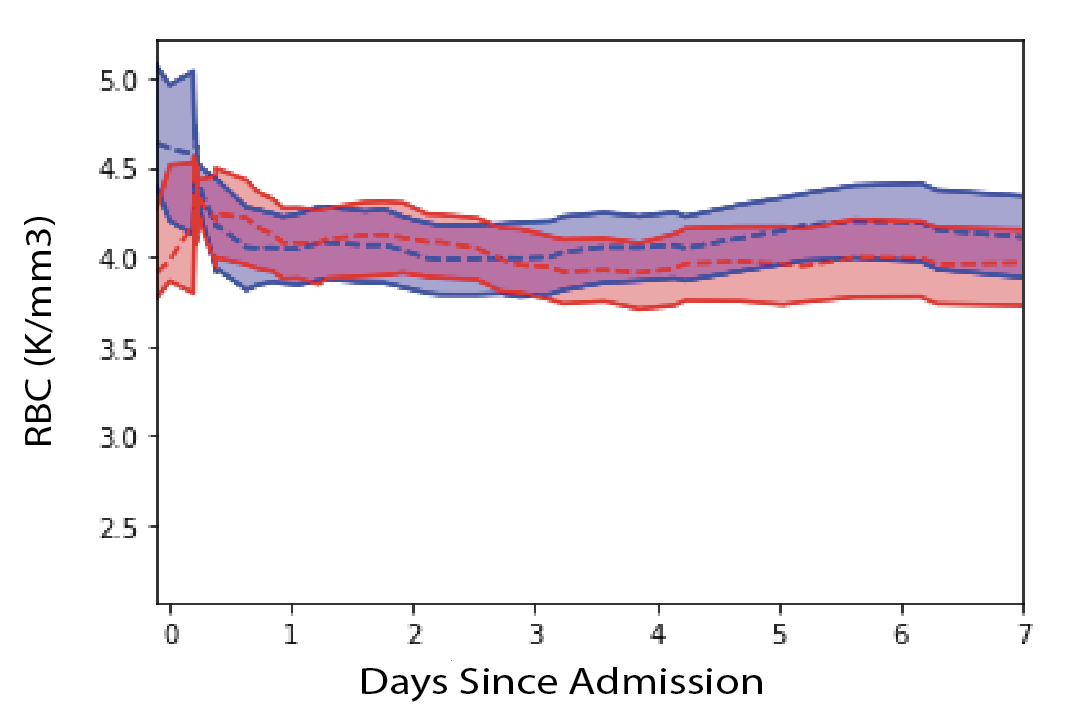}
\begin{center}
 
\par\end{center}%
\end{minipage}\hfill{}% 
\begin{minipage}[t]{0.23\columnwidth}%
\includegraphics[scale=0.24]{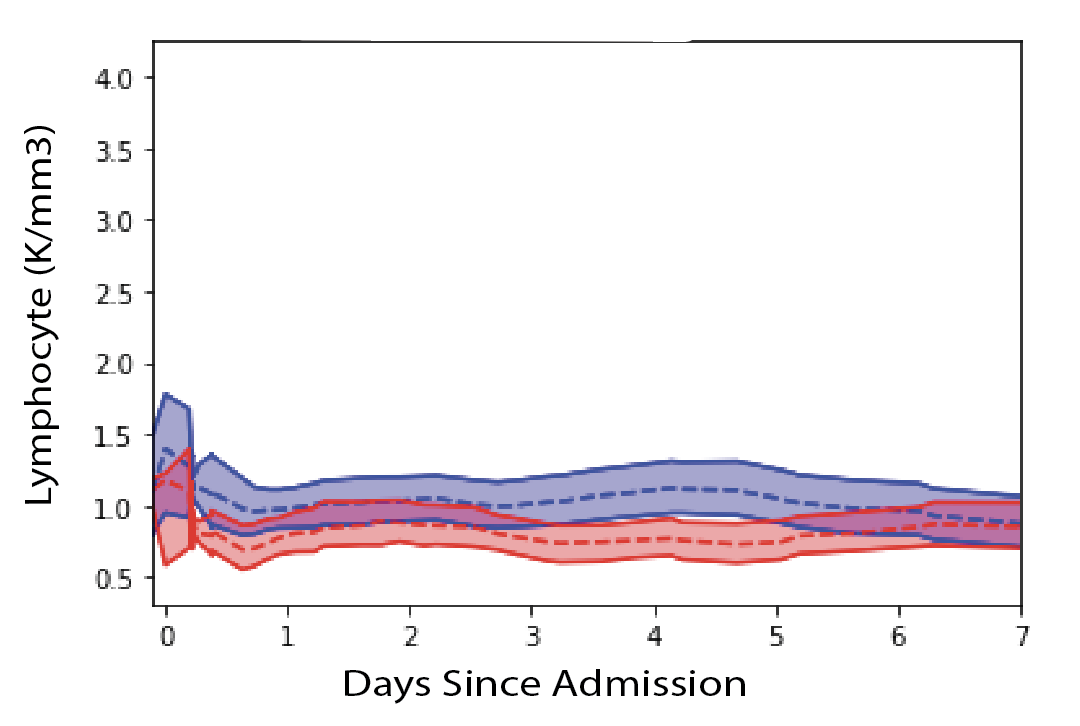}
\begin{center}
 
\par\end{center}%
\end{minipage}\hfill{}%
\begin{minipage}[t]{0.23\columnwidth}%
\includegraphics[scale=0.24]{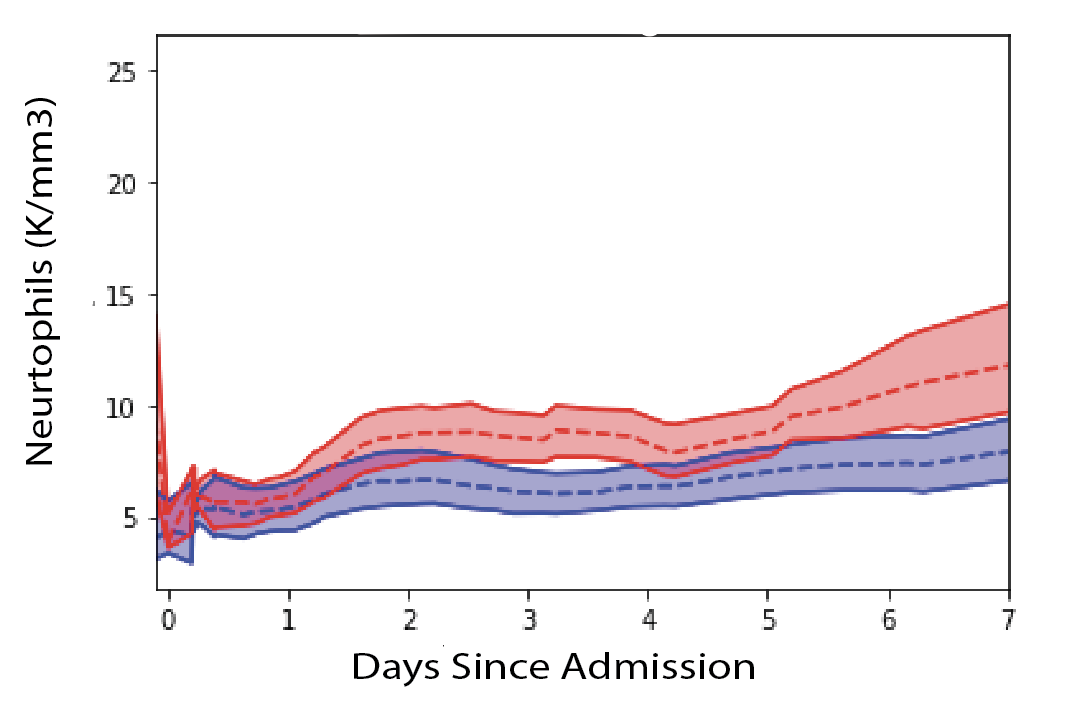}
\begin{center}
 
\par\end{center}%
\end{minipage}\hfill{}%
\begin{minipage}[t]{0.23\columnwidth}%
\includegraphics[scale=0.25]{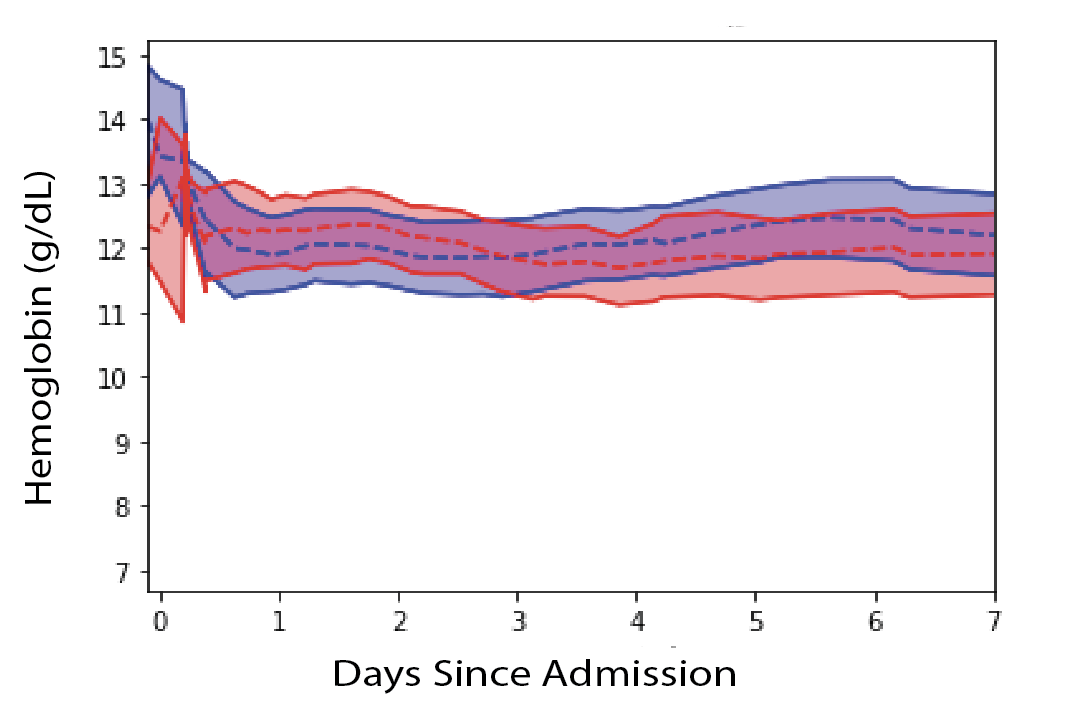}
\begin{center}
 
\par\end{center}%
\end{minipage}\hfill{}%
\begin{minipage}[t]{0.23\columnwidth}%
\includegraphics[scale=0.25]{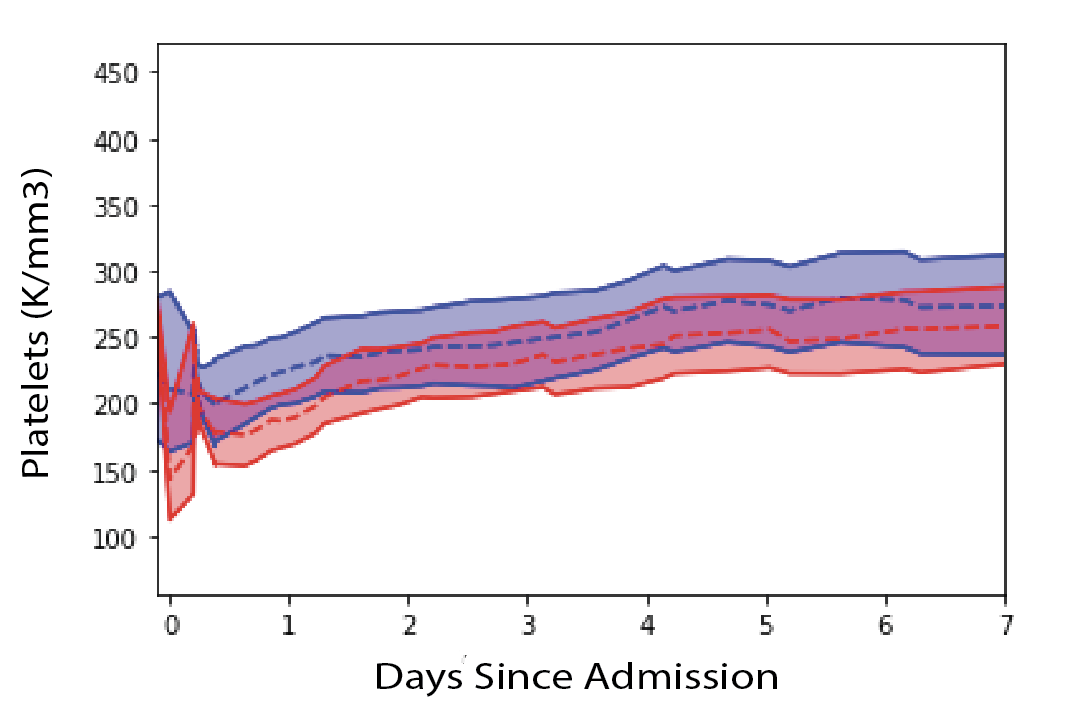}
\begin{center}
 
\par\end{center}%
\end{minipage}\hfill{}% 
\begin{minipage}[t]{0.23\columnwidth}%
\includegraphics[scale=0.25]{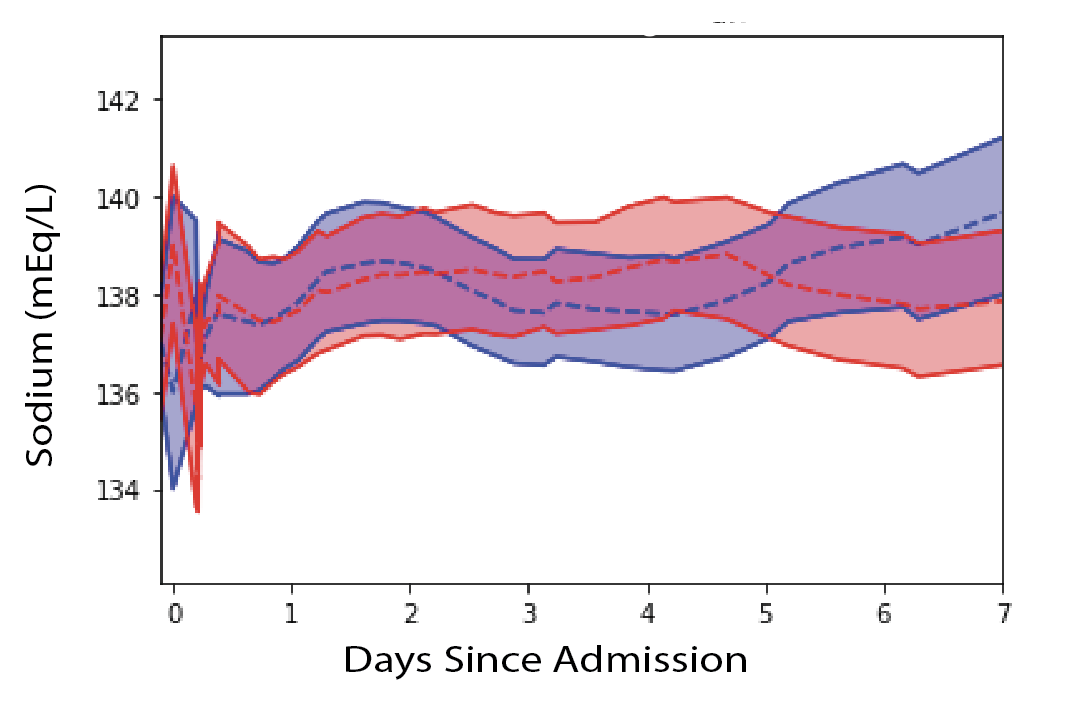}
\begin{center}
 
\par\end{center}%
\end{minipage}\hfill{}%

\begin{minipage}[t]{0.23\columnwidth}%
\includegraphics[scale=0.25]{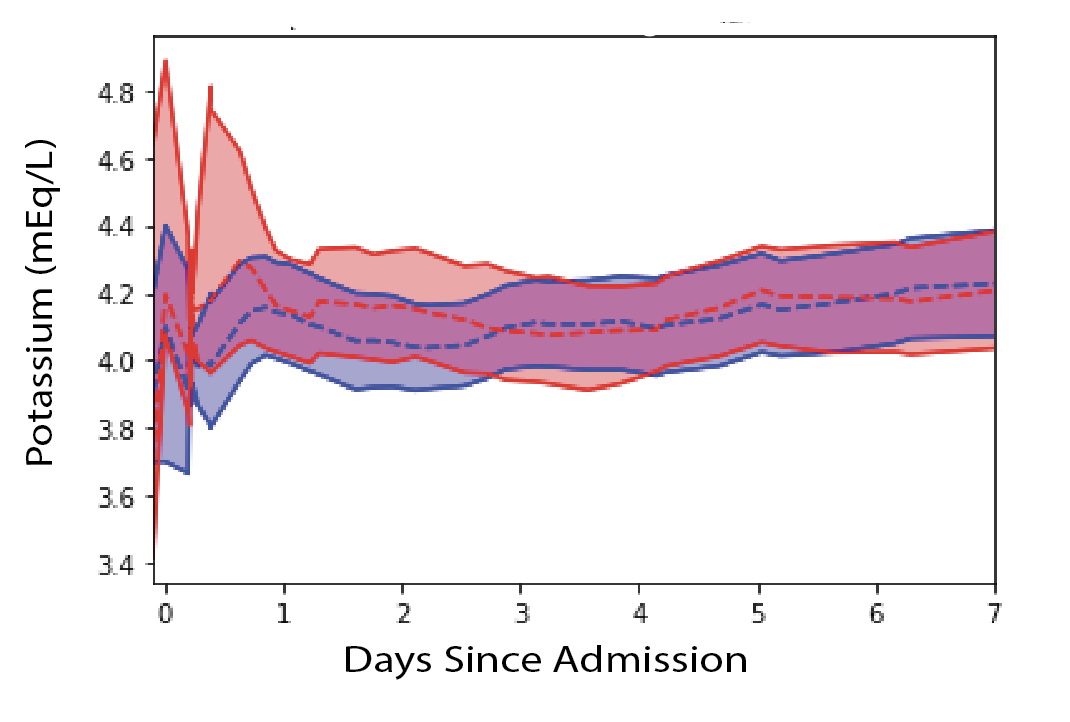}
\begin{center}
 
\par\end{center}%
\end{minipage}\hfill{}% 
\begin{minipage}[t]{0.23\columnwidth}%
\includegraphics[scale=0.25]{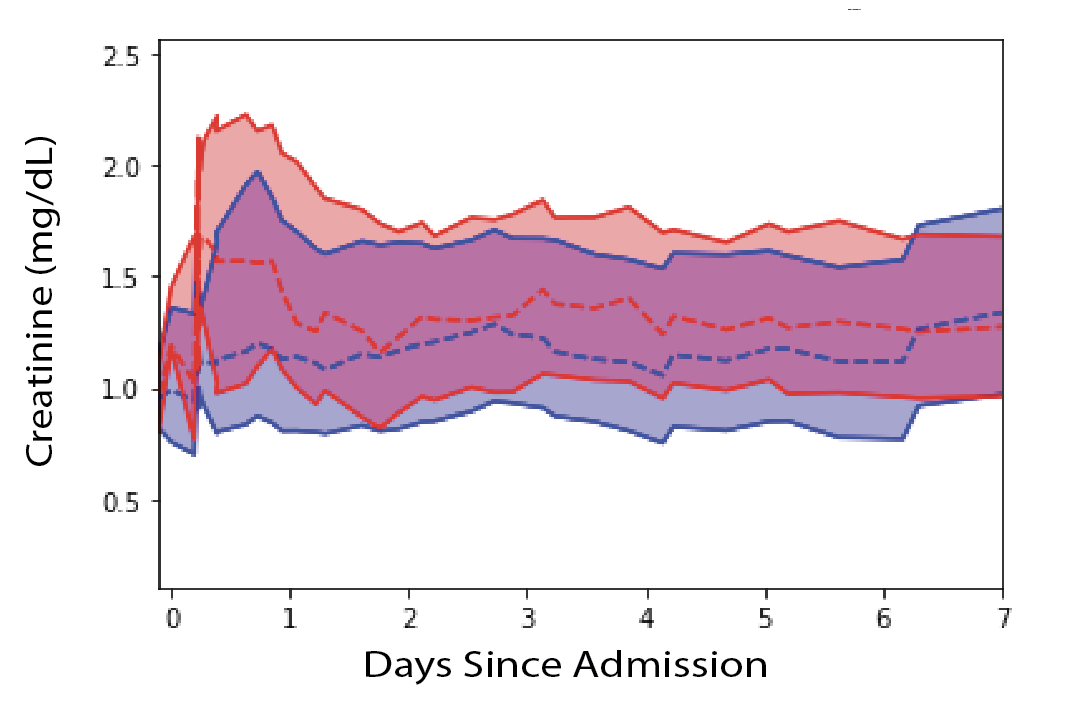}
\begin{center}
 
\par\end{center}%
\end{minipage}\hfill{}% 
\begin{minipage}[t]{0.23\columnwidth}%
\includegraphics[scale=0.25]{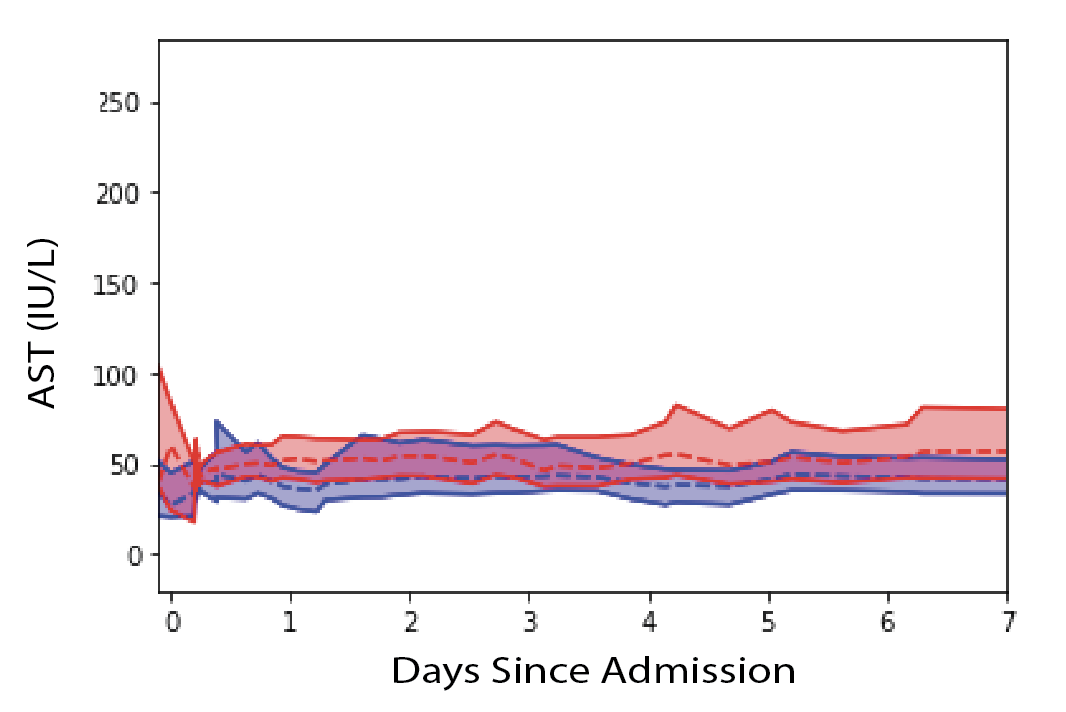}
\begin{center}
 
\par\end{center}%
\end{minipage}\hfill{}% 
\begin{minipage}[t]{0.23\columnwidth}%
\includegraphics[scale=0.25]{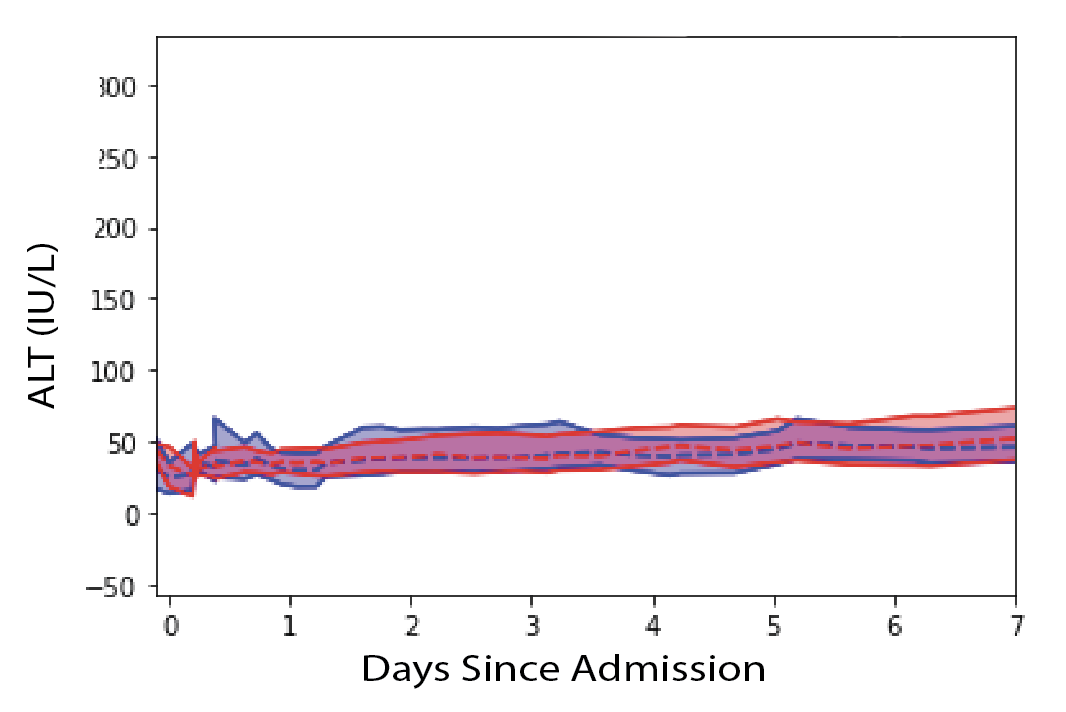}
\begin{center}
 
\par\end{center}%
\end{minipage}\hfill{}% 

\caption{IQR (filled) and median (dashed) of features (excluding age) included in our COVID severity prediction model after tSMOTE and smoothing has been applied.  Blue: patients with initial and WHO score $<5$.  Red: patients with initial WHO score $<5$ with max WHO score $>=5$. }
\label{fig:covFeats}
\end{figure}

We base these predictions on features which are readily available to clinicians upon admittance to the hospital. These include age,  vital signs (SpO$_2$, systolic/diastolic blood pressure,  temperature,  pulse), complete blood count (white/red blood cells, lymphocytes, neutrophils,  hemoglobin, platelets) and the basic metabolic panel (sodium, potassium, creatinine, alt, ast). To ensure the validity of tSMOTE, we took only patients who had these three sets of features measured within 1 hour before or after the others. This allows us to both ensure the disparate tests were taken at approximately at the same time, as well as ensuring tSMOTE can be readily applied without destroying correlations between the variables.  Once the samples have been fully imputed using tSMOTE, we apply a nonuniform Savitzky-Golay filter \cite{savgol} to smooth the trajectories before training the model. The ensemble of trajectories for the vital signs and labs after the smoothing has been applied are displayed in Fig.  \ref{fig:covFeats} as interquartile range (IQR) and median values.  

To train the model, we split our data using a 90-10 training-test split and z-normalize all of the data, using the mean and standard deviation of only the training set. We then use 10-fold cross-validation to train both the LSTM and logistic regression models simultaneously.  We partition our time interval into 100 time slices of varying length. Given the density of samples for times directly after admission, the first 46 points in
the time series correspond to the behavior over the first 12 hours of a patient's
stay, while the next 30 give us the behavior over the next seven days.  Once this is done, we use logistic regression on the final observation to classify into one ``moderate'' or ``severe'' COVID patients based on their max WHO score over that time. One can adjust the length of the input and output sequences to give any desired prediction model, although
the accuracy of the result will diminish with the size of the final time slices. We note that other classifiers can be used aside from logistic regression. Another obvious choice would be to have a dense neural network layer as the classifier. We only choose logistic regression since it is widely used in the medical community and offers an example of nice integration of our method with those which are commonly used.

We compare our model with tSMOTE data to classical logistic regression, random forest and XGBoost models with aggregated data.  Some error metrics for all three are shown in \ref{tab:aggs}. As you can see, the accuracy, and AUC  are all higher for our model with tSMOTE, showing a marked improvement over aggregated data for time forecasting.

\begin{comment}
We compare the performance of this particular algorithm with the tSMOTE procedure
against imputing missing observations with mean and median values within each time
slice. The summary statistics and confusion matrices are displayed in Table\ref{testResults}
and Fig. \ref{fig:cm}, respectively. We note that while mean and median imputation
perform well for training sets, it is largely due to the redundancy of the training
data. A majority of the samples have <5 observations initially, which means that
$95\%$ of the time series used to train the model is coming from imputed data. When
we impute and train in this way, the model will only recognize patient trajectories
which are extremely close to the mean or median trajectories. With tSMOTE on the other hand, the diversity
of training examples within each class is strengthened, leading to a more robust
model which can accurately characterize a larger number of examples. To illustrate
this, we train our model with each of the imputation techniques discussed above,
but evaluate it on samples which are only imputed with tSMOTE. This is to mimic
the actual diversity of trajectories present in real world patients. After this procedure,
we find that tSMOTE outperforms the mean and median imputation.
\end{comment}

\begin{table}[t]\label{tab:aggs}
\begin{centering}
\begin{tabular}{|c|c|c|c|}
\hline 
 & MSE (LSTM only) & AUC &  Accuracy\tabularnewline
\hline 
\hline 
LSTM/LR w/tSMOTE & 0.9144 & \textbf{0.9906}   & \textbf{0.9970}  \tabularnewline
\hline 
LR (aggregate) & N/A  & 0.7762 & 0.7383 \tabularnewline
\hline 
Random Forest (aggregate)& N/A  & 0.8036 & 0.7168 \tabularnewline
\hline
XGBoost (aggregate) & N/A & 0.8062 & 0.7226 \tabularnewline
\hline 
\end{tabular}
\par\end{centering}
\caption{AUC and Accuracy for our COVID prediction model, compared with three commonly used classifiers with aggregated time series data. }
\end{table}

\section*{Discussion}

Here we presented a novel imputation technique which can help turn sparse and irregular
time series data into uniform, albeit still irregular time series data. By slicing
our time interval into irregularly sized, non-overlapping bins, and by performing
linear interpolation along each feature direction independently, we are able to generate
new, synthetic observations which mimic those of the original data set, but are
different enough to capture the diversity of real world time series
data.  This procedure allowed us to both impute missing observations in their entirety
and impute individual features (when the variables are independent). To our knowledge this is the first time series analysis technique which allows one to do both of those things simultaneously. 

We illustrated our technique in a number of examples. First we showed how well we could reproduce a well known and well studied time series in two dimensions--the 2D uncoupled harmonic oscillator. This allowed for a controlled comparison where the shape of the time series is known a priori. Additionally, this is a natural setting for comparing tSMOTE against two obvious alternatives: imputing the mean and median of each time slice into missing observations. To make this comparison, we use an Encoder-Decoder LSTM model and a logistic regression model to distinguish between the trajectories of two 2D harmonic oscillators with different ratios of frequencies. There we saw a large improvement over the mean and median alternatives for classification.

Next, we applied tSMOTE to vital signs and lab results of COVID-19 positive patients to predict maixmum COVID severity after seven days based on the first twelve hours of patient data. We used the same model as in the 2D oscillator example, and again found a large improvement over models where time series data is aggregated.  Additionally, this model lets us easily chose the time frame which we wish to use as input data (e.g. first twelve hours of hospital stay) and output (e.g. seven days after admission). We also bypass the need for special RNN architectures which may be difficult for non-machine learning specialists to employ or interpret. 

\begin{comment}
However, given the homogeneity of these training samples, the variance of the classifiers trained on samples imputed with these methods will vanish, thereby maximizing bias leading to an underfit which leads to poor generalizable model which will miss important relations and fluctuations which may be important for classification.  Stated another way, when the mean and medians are imputed, most trajectories of a given class look the same and so the model can more accurately classify these trajectories.  However, when novel data needs to be classified, if the resulting trajectory contains fluctuations (which it undoubtedly will) these classifiers will perform poorly. When the model is trained on the mean or median imputation, and tested using completely synthetic data from tSMOTE, the accuracy slightly drops and the MSE increases. The accuracies are still comparable (all three are being >98$\%$), but the variance increases leading to a more balanced model. We find this to be preferable to a marginally more accurate model with a much higher bias-to-variance ratio. 
\end{comment}

There are a number of aspects of tSMOTE we wish to improve on future. Chief among them is a general strategy for choosing missing observations for imputation, such as a variational principle.  Additionally, we would like to be able to more faithfully represent the distribution of samples by carefully choosing missing observations which maximally preserves the initial  probability distribution. Both of these will allow us to apply our procedure to a wider class of data where predefined classes are not necessary for accurate imputation, and give more confidence in the analysis and interpretation performed on the imputed data set.

\section*{Online Methods}

\subsection*{Building tSMOTE}
Here we expand upon the steps involved in using tSMOTE presented in the main text.

\subsubsection*{Constructing and Assigning the Time Slices}\label{subsec:Constructing-and-Assigning}

We start by constructing the time slices. For classification problems, this should
be done before the data has been sorted into classification classes to ensure the
number of time slices, as well as their lengths, are equal across all the classes. Once the slices have been made, the tSMOTE procedure should be performed within each class, so that the resulting observations look like other observations in the class. The
pseudocode is presented below as Algorithm \ref{alg:Assign}.

First, we must pick a relevant starting point. This is highly dependent on the problem
at hand, but should be computable, or at least inferable, from the data directly.
Next, we must pick a maximum endpoint. The most straight forward approach is to chose
the following:

\begin{eqnarray*}
t_{max} & = & \max_{classes}\left(\max_{i}t_{m_{i}}^{i}\right)\\
t_{min} & = & \min_{classes}\left(\min_{i}t_{1}^{i}\right)
\end{eqnarray*}
although one could pick a smaller value for $t_{max}$ if it suits the problem (for instance if the largest time window is parametrically larger than the
next largest one).
Similarly, one may wish to choose a different start time, such as symptom onset (for
medical diagnosis/prognosis problems). We caution against picking a larger $t_{max}$
since there will be no data present in the time slices larger than $t_{max}$. Once
these parameters have been found, we compute the elapsed time since the given start
time as an array $\Delta T$. 

Once we have obtained $\Delta T$, we tag each observation with both it's $\mu$ and
$i$ indices and flatten the array. We then sort the resulting flat list by the
observation and partition the list such that there exist $n_{T}$ slices ($n_{T}$
being specified by the user) with an equal number of observations in each slice. We
can then use the tags to go back to the original data set and assign to each sample $\bm{x}_{i}(t)$ a vector of integers which represent the time slices to which this sample belongs.  This leads to the following definition of a time slice:
\begin{definition}
\label{def:The-n-time}The n$^{th}$ \textbf{\emph{time slice }}$\delta T^{n}$ is
defined as $\delta T^{n}\equiv[\sum_{i=1}^{n-1}\delta t_{i},\sum_{i=1}^{n}\delta t_{i})$
where $\delta t_{i}$ is the size of the $i^{th}$ slice calculated from the initial
partitioning procedure.
\end{definition}
In this definition we take the lower bound to be closed so that the sample from
which $t_{min}$ was obtained belongs to the first time slice, while we leave the
upper bound open to ensure a given observation does not exist in more than one time
slice.

\subsubsection*{Data Generation}\label{subsec:Data-Generation-and}

After the time slices have been assigned and sorted, we are ready to employ the SMOTE algorithm. To do so, we rephrase this imputation problem into
a class imbalance problem: 
\begin{definition}
\label{def:The--component}The $k^{th}$ component of the sample $\bm{x}_{i}(t)$\emph{
}\textbf{\emph{belongs to $\delta T^{n}$}} if i) $\left[\bm{x}_{i}(t_{\mu}^{i})\right]_{k}$
is not NULL and ii) $t_{\mu}^{i}-t_{min}\in\delta T^{n}$ for at least one $\mu\in\left\{ 1,2,...,m_{i}\right\} .$
\end{definition}
In other words, the $k^{th}$ component must have been measured in the appropriate
time range for it to belong to the slice. 
\begin{definition}
\label{def:A-data-point}A sample $\bm{x}_{i}(t)$\emph{ }\textbf{\emph{belongs to
$\delta T^{n}$}} if all of its components belong to $\delta T^{n}$. If this condition
is met for more than one $\mu$, we call the sample $\bm{x}_{i}(t)$ \textbf{\emph{degenerate
in}} $\delta T^{n}$ or \textbf{\emph{degenerate in n}}.
\end{definition}
The idea now is to ``balance'' these classes in such a way that each sample
belongs to each time slice. This means two things: 1) ensure each component $\bm{x}_{i}(t_{\mu}^{i})$
belongs to the time slice, and 2) ensure each $\bm{x}_{i}(t)$ belongs to every time
slice. Before we do this, however, we must sort the samples into their classification
classes to ensure we are drawing synthetic data from samples which belong to
the same class. This is absolutely crucial for tSMOTE to approximate real trajectories in a manner that preserves the distinct behavior between the different classes. After this has been done, we generate a surplus of synthetic samples for each class
using SMOTE within each time slice. 

To address problems 1), 2) and 4) simultaneously we slightly
generalize the SMOTE algorithm by performing the data generation along each feature direction
individually. This is not possible using the original SMOTE procedure, since it necessarily
requires all the individual features to be present. In particular, we first generate
individual \emph{features }in the following way: given an individual feature of a
particular sample in the first time slice, $[\bm{x}(\delta T^{1})]_{k}$ , we first
find its nearest neighbor along the $k^{th}$ direction in that slice $[\bm{y}(\delta T^{1})]_{k}$,
and generate a new sample by computing: 
\[
[\bm{x}_{new}(\delta T^{1})]_{k}=[\bm{x}(\delta T^{1})]_{k}+\lambda([\bm{y}(\delta T^{1})]_{k}-[\bm{x}(\delta T^{1})]_{k})
\]
where $\lambda\in(0,1)$ is a random number. It is interesting to note that the nearest
neighbors along each direction may not be the same sample. As a result, we are
not simply picking a point on a straight line connecting two existing points, as
in SMOTE, but instead generating a new sample at an arbitrary location within the nearest neighborhood. This procedure works well for samples in which all the components are presents, i.e. data sets which do not satisfy issue 4). However, as we will discuss below, this will destroy correlations between features when samples have individual features missing.  If correlations between features is small or zero, then this procedure still works well.  The pseudocode for this portion of the algorithm can be found below as Algorithm \ref{alg:generate}.

\subsubsection*{Imputation}

Once we have generated our synthetic time slice data, we can choose new observations from
these time slices and impute them into the samples which do not belong
to that time slice. First, we go through each sample and replace their null values
with the appropriate values from a randomly selected synthetic sample. Then,
we fill out the full sample's trajectory by randomly sampling feature vectors
from our synthetically generated samples, within the appropriate class and time
slice. To define this latter process, we reshape our initial samples from an array of shape $(m_{i},n_{F})$
to an an array of shape $(n_{T},n_{F})$
leaving nulls when the sample does not belong to a given slice. Specifically, we
have 
\begin{eqnarray*}
I:V^{m_{i}} & \to & V^{n_{T}}\\
\bm{x}(t_{i}) & \mapsto & \bm{x}(\delta T^{j})\:\text{if }t_{i}\in\delta T^{j}
\end{eqnarray*}
where $V$ is the $n_F$-dimensional vector space on which our data is defined, likely some combination
of $\mathbb{R}$ and $\{0,1\}$. By the nature of this problem, this map is not surjective
(unless we are extremely lucky). However, it is also not injective, since a data
point might be degenerate in some subset of time slices. This is bad because we are
implicitly trying to construct an injective map, so that we may obtain one single,
unique trajectory for each sample. To deal with this, we take the average of
degenerate samples in a given slice. This ensures our imputation map is injective. Once this map has been applied, we randomly sample from our surplus of synthetic
data to replace the nulls with these synthetic samples. The pseudocode for this
portion of our method is presented below as Algorithm 3. 

To illustrate the imputation step, consider the following example.
\begin{example}
Consider a sample $\bm{x}(t)=(\bm{x}(t_{1}),\bm{x}(t_{2}),...,\bm{x}(t_{m}))$ where each $\bm{x}(t_{i})$
contains no null components. Suppose we find that $t_{1}\in\delta T^{1}$ while $t_{2}\in\delta T^{4}$.
At the moment, sample $\bm{x}(t)$ \emph{does not belong to} $\delta T^{2}$ and $\delta T^{3}$.
By the imputation map above, we obtain 

\[
(\bm{x}(t_{1}),\bm{x}(t_{2}),...,\bm{x}(t_{m})\mapsto(\bm{x}(\delta T^{1}),\,\boldsymbol{\cdot}\,,\,\boldsymbol{\cdot}\,,\bm{x}(\delta T^{4}),...,\bm{x}(\delta T^{n_{T}})).
\]

Next, we find the position of the NULL entries, and draw a random sample from the
corresponding synthetically generated time slices. For slices $\delta T^{2}$ and $\delta T^{3}$
our sampling gives $\bm{y}(\delta T^{2})$ and $\bm{z}(\delta T^{3})$ . The final trajectory of
our sample will then look like 

\[
\bm{x}_{fin}(t)=(\bm{x}(\delta T^{1}),\bm{y}(\delta T^{2}),\bm{z}(\delta T^{3}),\bm{x}(\delta T^{4}),...,\bm{x}(\delta T^{n_{T}})).
\]
\end{example}
There is one small subtlety that needs to be addressed in the process above. There
could be some features that do not change over time. These could be fixed physical
characteristics (e.g. height), preexisting conditions (e.g. asthma), long term medications,
etc.... In this case, we would need to be sure that the resulting synthetic sample fixes these particular features. This can be achieved in one of two way: do
SMOTE on the time varying features only or simply replace these fixed features with
the correct ones after drawing new samples. 

\subsubsection*{Smoothing}
\begin{figure}[H]
\begin{minipage}[t]{0.33\columnwidth}%
\includegraphics[scale=0.3]{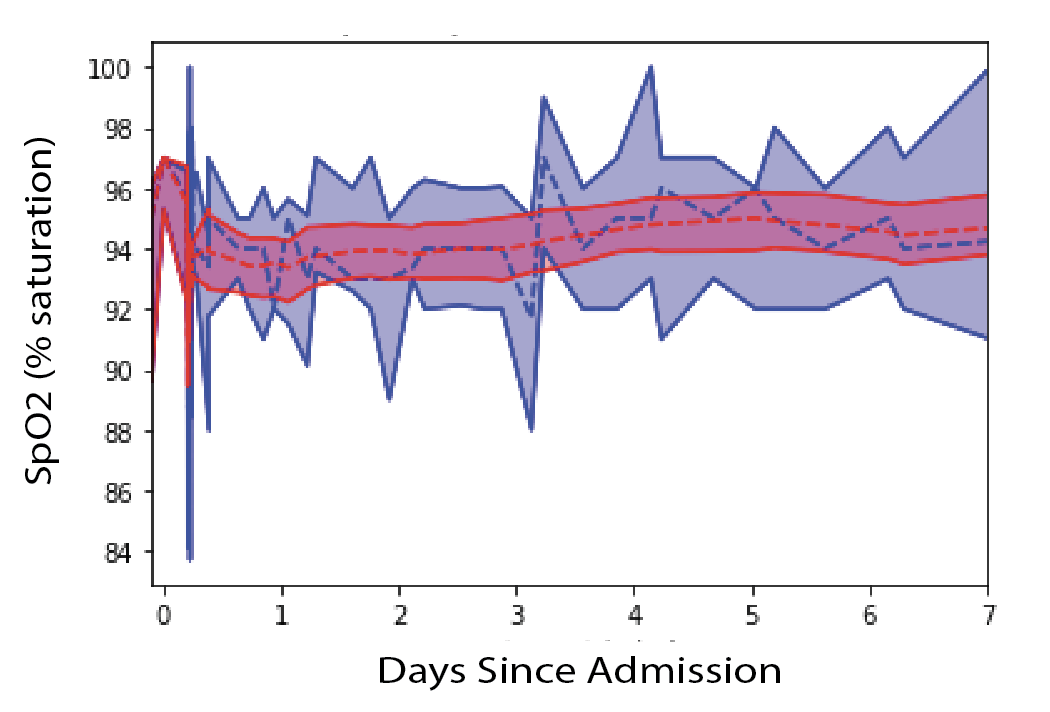}
\begin{center}
 
\par\end{center}%
\end{minipage}\hfill{}%
\begin{minipage}[t]{0.33\columnwidth}%
\includegraphics[scale=0.3]{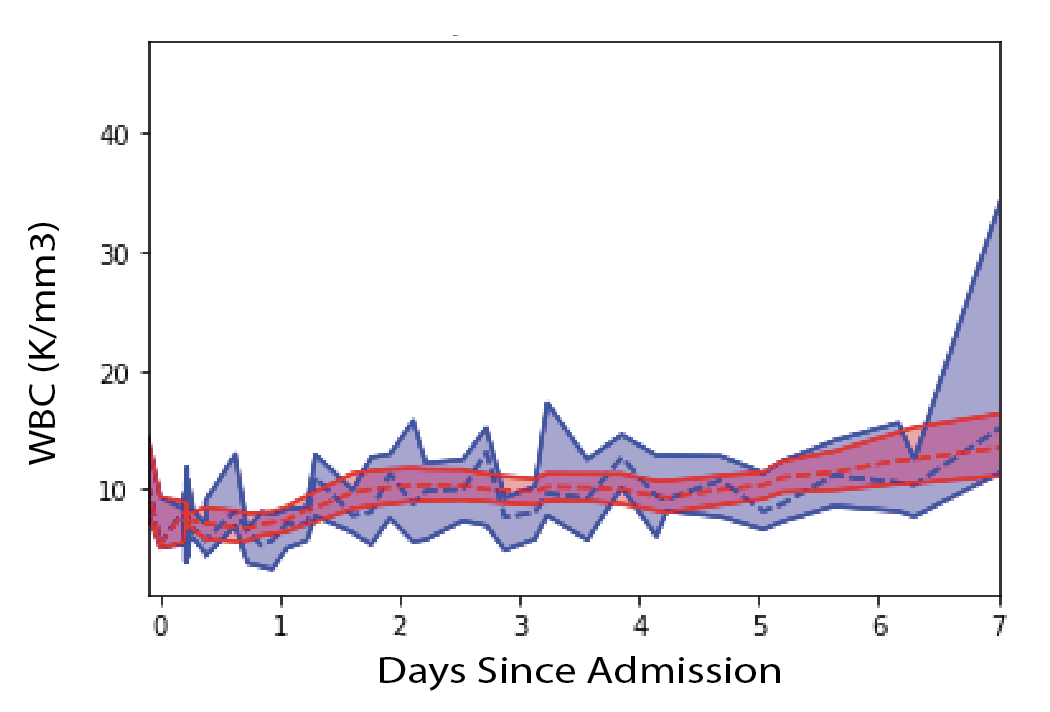}
\begin{center}
 
\par\end{center}%
\end{minipage}\hfill{}%
\begin{minipage}[t]{0.33\columnwidth}%
\includegraphics[scale=0.3]{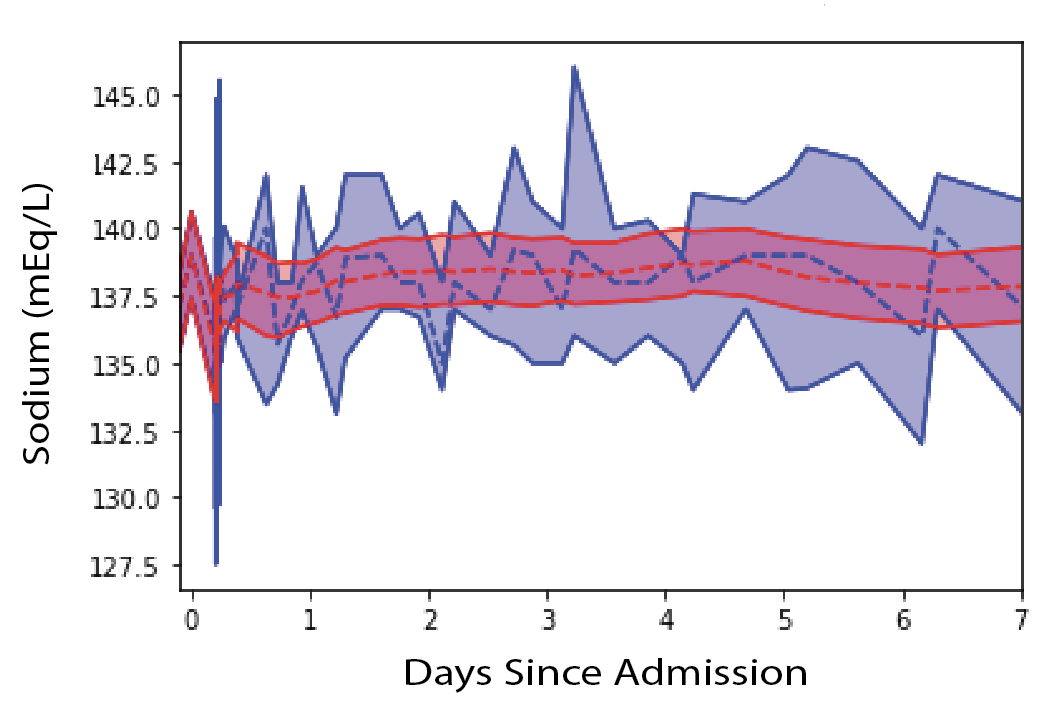}
\begin{center}
 
\par\end{center}%
\end{minipage}\hfill{}% 
\caption{IQR (filled) and median (dashed) of selected features before (blue) and after (red) a non-uniform Savitzky-Golay filter has been applied.}\label{smoothing}

\end{figure}

As a final step, we apply a smoothing filter to the fully imputed time series. For this we use a non-uniform Savitzky-Golay filter \cite{savgol}, which works by fitting a low degree polynomial to successive subsets of the data and replacing the midpoint by value of the polynomial at that time point.  Specifically, we use a window size of twenty-five and a polynomial of order five.  We note that other smoothing kernels could also work here, such as Gaussian or kernel average smoothing.  The main purpose is to reduce high frequency in the resulting imputed time series without distorting the signal. This step is not completely necessary, but it improves performance on models such as those discussed in the main text.  As an example,  Fig.  \ref{smoothing} displays some of the features used in our COVID-19 prediction model before and after smoothing was applied to each trajectory. 

\subsection*{Improvement Over Mean and Median Time Slice Imputation}
Let's consider the case of mean imputation (the same calculation applies to median imputation without significant alteration) and see why tSMOTE is preferred. The mean of each time slice is not significantly altered by any of these imputation procedures (see below for details in the tSMOTE case), but let us consider the second moments of the data. Prior to imputation each time slice has $n$ points in it and the variance of the slice is given by 
\begin{equation}
\sigma^2=\frac{1}{n}\sum_{i=1}^{n} \left( \bm{x}_i - \bar{\bm{x}} \right)^2.
\end{equation}
After imputing the mean of the time slice for each sample which are not in the slice,  the time slice now contains all $N$ samples. The new variance is 

\begin{align*}
\tilde{ \sigma }^2 &=\frac{1}{N}\sum_{i=1}^{N} \left( \bm{x}_i - \bar{\bm{x}} \right)^2 \\
& = \frac{1}{N}\sum_{i=1}^{n} \left( \bm{x}_i - \bar{\bm{x}} \right)^2 \\
& = \frac{n}{N}\sigma^2
\end{align*}
and since $n=N/n_T$, with $n_T$ being the number of time slices, we have 
\begin{equation}\label{varBad}
\tilde{\sigma}^2=\frac{1}{n_T}\sigma^2
\end{equation}
which is a significant reduction for most reasonable values of $n_T$. On the other hand, if we instead added samples from our set of tSMOTE generated data, we would find:
\begin{align*}
\tilde{\sigma}^2 &= \frac{1}{N}\sum_{i=1}^{N} \left( \bm{x}_i - \bar{\bm{x}} \right)^2 \\
&= \frac{1}{N}\sum_{i=1}^{n}\left( \bm{x}_i - \bar{\bm{x}} \right)^2 + \frac{1}{N}\sum_{i=1}^{N-n}\left(\bm{x}^{tS}_i-\bar{\bm{x}}^{tS}_i\right) \\
&= \frac{1}{n_T}\sum_{i=1}^{n}\left( \bm{x}_i - \bar{\bm{x}} \right)^2 + \left(1-\frac{1}{n_T} \right)\sum_{i=1}^{N-n}\left(\bm{x}^{tS}_i-\bar{\bm{x}}^{tS}_i\right)^2  \\ 
&=\frac{1}{n_T}\sigma^2+\left(1-\frac{1}{n_T}\right)\sigma_{tS}^2.
\end{align*}
As we will show below, the variance of the tSMOTE generated data $\sigma^2_{tS}$, while not exactly equal to the variance of the original set, is of comparable magnitude. We can write $\sigma_{tS}^2=\sigma^2-\epsilon$, for some error term $\epsilon$.\footnote{The minus sign here is meant to reflect the analysis performed in the appendix of the initial SMOTE paper \cite{smote}, where simulations show that, in general, the variance of SMOTE generated data is smaller than the variance of the initial data set, and approaches $2/3$ of the initial data when the number of features is large.  Additionally we show below that the sample variance can be written as $\sigma^2-\epsilon$ and that $\bb{E}(\sigma_{tS}^2)=(2/3)\bb{E}(\sigma^2)$ when the observations are i.i.d.}.  Plugging this in, we have
\begin{equation}\label{varGood}
\tilde{\sigma}^2=\sigma^2-\left(1-\frac{1}{n_T}\right)\epsilon
\end{equation}
So while the in-slice variance of the imputed data is not necessarily equal to the initial data, it is a marked improvement over eq. \eqref{varBad}. 

Preserving the variance of imputed data is an important part of an imputation process for many reasons. Firstly, it is important for discriminant analyses such as classification because classifiers require a large range of training examples to accurately assign new samples to their predicted class. If the variance of the data is too small, the classifier will only be able to accurate classify a very small range of samples. On the other hand, if it is too big, the classification regions may overlap and it will be difficult to discriminate between classes.  Given this view, it may be beneficial to slightly decrease the variance, as shown in eq. \eqref{varGood}. Second, the variance of the data may itself contain pertinent information. If the goal of the analysis is to characterize the probability distribution, it would be important to preserve the moments of the distribution, otherwise the inference will not be accurate.  Third, the variance of, the trajectory itself may contain important information. For instance, if a patient has a large variance of their BMI, or blood pressure, it could be indication of increased risk for certain diseases\footnote{This is not exactly applicable to the current implementation of tSMOTE, since the individual trajectories are not yet reflective of the actual trajectories, but will be important for future improvements on the current algorithm.}.

\subsection*{Limitations}

In addition to the usual considerations about the validity of statistical methods
(more data=better!), tSMOTE will perform best if i) $n_{D}\cdot\sum_{i}m_{i}\gg n_{T}$
and ii) $t_{\mu}^{i}$ is generated via a Poisson process. Condition i) states that
the total number of observations across each class must be sufficiently large in order
to trust the accuracy of the resulting imputation. For it to even work in the first place we would need at least two points in each time slice, leading to the lower bound $n_D \ge 2n_T$. Still, to saturate this bound we would need the samples to be evenly distributed among the time slices, which is not always guaranteed.  In order to
ensure each time slice contains the minimum allowable number of samples, and
to allow for the irregularity of the time series data, we must have $n_{D}\cdot\sum_{i}m_{i}\gg n_{T}$.
Of course, one could always decrease the number of time slice present, but ideally
one would like to have as many time slices as allowed by the data. 

Condition ii) says that each observation should be randomly picked from a uniform interval, defined by our start and end time. This condition will ensure that the resulting observations
are reasonably well distributed across the entire interval, so that each time slice
will be approximately the same length. Moreover, this will help to minimize degeneracies which requires averaging over the values in a time slice
of a given sample, and replace the missing observations with synthetic data. This
is not ideal, since we should be using as much of the original data as possible.
By averaging, we are throwing away important information about the data in that slice.
Additionally, having equal-sized time slices allows one to more consistently reconstruct
the true trajectories of the time slice data alone. Together with condition i) we
could choose sufficiently fine-grained time slices which will allow for the best
possible trajectory reconstruction.

The primary advantage to interpolating along each axis individually is to utilize
the totality of the data, including samples which have null entries for features
at sporadic observations. However, there may be cases where the nearest neighbors,
as defined by the Euclidean distance between the samples, actually gives a worse
approximation for a point in the true data subspace in the neighborhood of $\bm{x}_{i}$
compared to our method. For instance, consider a case where our data lives on a Riemannian
manifold embedded in $\mathbb{R}^{n_{f}}$ with the curvature along one direction much
larger than the curvature along the others. If the true nearest neighbor has a large
projection along the highly curved direction and a small projection along the other
directions, then simple interpolation to that point might not yield a point which
well approximates the manifold in that neighborhood. However, if there is another
point which has a larger Euclidean distance, but which has a smaller projection along
the highly curved direction, first projecting and then interpolating will yield a
point which better approximates the true data manifold. We should note that we do
not yet have a rigorous proof of this fact, and so the claim should only serve as
a heuristic explanation into a potential improvement.  

The biggest room for improvement in tSMOTE comes in the imputation step. Currently our imputation step relies on the prior stratification of the data rather than on some selection criterion. For example, we split our COVID cohort into moderate and severe cases so that we may impute observations which were made from synthetic data within that class. This works well for training classifiers or increasing sample sizes, but does not allow for a good interpretation of the individual trajectories themselves.  Nevertheless, this is sufficient for training an RNN, or analyzing an ensemble of trajectories as opposed to individual ones. We hope to improve this step so more interpretable conclusions can be drawn from the imputed trajectories. 

\subsection*{Pseudocode}

Here we present the pseudocode for the algorithms invloved in tSMOTE.

Algorithm \ref{alg:Assign} contains the pseudocode for our initial assignment of the time slices. Algorithm \ref{alg:generate} describes our slightly modified SMOTE algorithm, which allows us to use samples where individual nulls are present. Finally, algorithm \ref{alg: impute} is how we impute the synthetic data into the initial data. 

\begin{algorithm}[h]
INPUT~$T=[[t_{1}^{1},t_{2}^{1},...,t_{m_{1}}^{1}],[t_{1}^{2},t_{2}^{2},...,t_{m_{2}}^{2}],...,[t_{1}^{n_{D}},t_{2}^{n_{D}},...,t_{m_{n_{D}}}^{n_{D}}]]$~, $N_{S}$,~$t_{min},\,t_{max}$

COMPUTE~$\Delta T':=[[t_{1}^{1}-t_{min},1,1],...,[t_{m_{1}}^{1}-t_{min},1,m_{1}],...,[t_{m_{1}}^{n_{D}}-t_{min},n_{D},m_{n_{D}}]]$~

SORT~$\Delta T'$~by~first~element

PARTITION~$\Delta T'$~into~$N_{s}$~slices~of~equal~size

COMPUTE~$\bm{dt}=$size~of~each~slice

slices={[}{[}{]},...,{[}{]}{]}~(empty~list~same~shape~as~$T$)

FOR~i~IN~range($|\Delta T'|$):~

\qquad FOR~j~in~range(|$\Delta T'[i]$|):~

\qquad \qquad label=$\Delta T'[i][j]$

\qquad \qquad EXTEND~slices{[}i{]}~BY~label+1

OUTPUT~slices,~$\bm{dt}$~

\caption{Procedure for Generating and Assigning Time Slices}
\label{alg:Assign}
\end{algorithm}

\begin{algorithm}[h]

INPUT~slice={[}$\bm{x}_{1},\bm{x}_{2},...,\bm{x}_{s}]$

synSlices={[}{]}

FOR~i~IN~range($N_{feats})$:~

\qquad feats={[}{]}

\qquad FOR~x~IN~slice:~

\qquad \qquad APPEND~x{[}i{]}~TO~feats

\qquad DROP~NULLS~FROM~feats

\qquad FOR~$z$~IN~feats:~

\qquad \qquad FIND~k-NearestNeighbors(z)

\qquad \qquad FOR~$y$~IN~k-NearestNeighbors(z):~

\qquad \qquad \qquad$\lambda$:=randomReal(0,1)

\qquad \qquad \qquad$z_{new}=z+\lambda(y-z)$

\qquad \qquad APPEND~$z_{new}$~to~synSlice

$T_{syn}$={[}{]}

FOR~j~IN~range($N_{points}$):~

\qquad vec={[}{]}

\qquad FOR~k~IN~range($N_{feats}$):~

\qquad \qquad APPEND~synSlic{[}j{]}{[}k{]}~TO~vec

\qquad APPEND~vec~TO~synSliceFin

OUTPUT~synSliceFin

\caption{Rough Implementation of SMOTE {[}ref{]} Within a Single Time Slice}
\label{alg:generate}
\end{algorithm}

\begin{algorithm}[h]

INPUT~$\mathbf{X}=[[\bm{x}_{1}(t_{1}^{1}),...,\bm{x}_{1}(t_{m_{1}}^{1})],...,[\bm{x}_{n_{D}}(t_{1}^{n_{D}}),...,\bm{x}\bm{x}_{n_{D}}(t_{m_{n_{D}}}^{n_{D}})]]$,~$T_{slices}$,~synSlices,~nFix

$X_{out}=${[}{]}

FOR~i~IN~range(|$\bm{X}$|):~

\qquad REPLACE~Nulls~in~$X[i]$

\qquad list={[}{[}{]},...,{[}{]}{]}~(list~of~$n_{T}$~empty~lists)~

\qquad FOR~j~IN~range(|$\bm{X}[i]$|):~

\qquad \qquad $t=T_{slices}[i][j]$

\qquad \qquad APPEND~$\bm{X}[i][j]$~to~list{[}$t-1${]}

\qquad FOR~$k$~IN~range(|list|):~

\qquad\qquad IF~list{[}$k${]}~IS~empty:~

\qquad \qquad \qquad C=randomChoice(synSlices{[}k{]})

\qquad \qquad \qquad IF~nFix>0:~

\qquad \qquad \qquad \qquad FIX~first~nFix~(time~independent)~features~elements~of~C~~

\qquad \qquad \qquad \qquad  EXTEND~C~to~list{[}$k${]}

\qquad \qquad \qquad ELSE~IF~|list{[}$k${]}|>1:~

\qquad \qquad \qquad \qquad AVERAGE~OVER~list{[}$k${]}

\qquad \qquad  APPEND~list~TO~$X_{out}$

OUTPUT~$X_{out}$

\caption{Implementation of The Imputation Map}
\label{alg: impute}
\end{algorithm}

\subsection*{Theoretical Properties of SMOTE}

Here we calculate some theoretical properties of tSMOTE within each slice. Similar calculations are done in \cite{smote}. Here we provide explicit formulas for the sample mean and covariance, and show that they reduce to the calculations in \cite{smote} when taking the expected value. Before we do this, we must introduce some new notation. Throughout, we consider operations only a single feature for notational ease.  Additionally, we assume all of our data points $x^{\mu}$ are i.i.d. 

We start with synthetic data of the form
\begin{equation}\label{smoteForm}
x_{new}=x_{old}+\lambda(y_{nn}-x_{old})
\end{equation}
where $y_{nn}$ is a nearest neighbor of $x_{old}$. Before we start, we need a better way to format this equation so that it can be easily manipulated. First, let superscript greek letters index the data points like $x^{\mu}$. Next, let $\Omega(x^{\mu})$ denote the set of K nearest neighbors of $x^{\mu}$. We then index elements of this nearest neighbor set with capital Latin letters such as I: $x^{I}\in\Omega(x^{\mu})$ . The above equation can be then written as 
\begin{equation}
x_{new}=x^{\mu}+\lambda^{\mu I}(x^{I}-x^{\mu})
\end{equation}
where we have the vector $ \lambda^{\mu I}$ is a random variable drawn from an arbitrary distribution over the unit interval.  

In this form, we can attempt to write our new data set as a matrix equation. To do this, we have to form a $D(K+1)$ vector of each data point and it's nearest neighbors. We do this by sequentially putting in a data point, followed by it's $K$ nearest neighbors, then another data point and it's $K$ nearest neighbors and so on. Let us call this data vector of data points and their nearest neighbors $\bm{X}$. This is an extended data matrix where each data point is duplicated $k_{\mu}+1$ times. We are then interested in the generation of $D\cdot M$ ( with $M\ge K$) new synthetic data points $\bm{Y}$. We can write the new data points as the following matrix equation 
\begin{equation}\label{eq:matForm}
\bm{Y}=\bm{\Lambda}\bm{X}
\end{equation}
where $\bm{\Lambda}$ is a matrix of size $D\cdot M\times D\cdot(K+1)$ and is of the form 

\begin{equation}\label{eq:blockDiag}
\Lambda=\left[\begin{array}{ccccc}
\bm{\Lambda^{1}}\\
 & \bm{\Lambda^{2}}\\
 &  & \ddots\\
 &  &  & \bm{\Lambda^{D-1}}\\
 &  &  &  & \bm{\Lambda^{D}}
\end{array}\right]
\end{equation}

where each individual $\bm{\Lambda^{\mu}}$ is an $M\times(K+1)$ matrix which acts on $x^{\mu}$ and it's neighbors in $\Omega(x^{\mu})$.  Each $\bm{\Lambda^{\mu}}$ is of the form
\begin{equation}\label{eq:explicitLambda}
\Lambda^{\mu}	=\left[\begin{array}{ccccc}
1-\lambda_{1}^{\mu1} &\lambda_{1}^{\mu1} & 0 & \cdots & 0\\
\vdots & \vdots & \vdots & \vdots & \vdots\\
1-\lambda_{R}^{\mu1} & \lambda_{R}^{\mu1}\\
1-\lambda_{1}^{\mu2} & 0 & \lambda_{1}^{\mu2} & \cdots & 0\\
\vdots & \vdots & \vdots & \ddots & \vdots\\
1-\lambda_{R}^{\mu K} & 0 & 0 & 0 & \lambda_{R}^{\mu K}
\end{array}\right].
\end{equation}

We now want to calculate the mean of $\bm{Y}$ and find its relation to the mean of $\bm{X}$. To make things easier for now, we will take $R=1$. In component notation, we can write eq. \eqref{matForm}as 
\begin{equation}\label{compForm}
Y^{\alpha}=\Lambda^{\alpha \beta} X^{\beta}
\end{equation}
where the Einstein summation convention is used. 

\subsection*{Mean}

Lets start by calculating the first moment of the elements of the vector $Y$, with $R=1$. This makes $\Lambda$ a $DK\times D(K+1)$ matrix. We have 
\begin{equation}\label{eq:startMean}
\begin{split}
\sum_{\alpha}Y^{\alpha} &=\sum_{\alpha}\Lambda^{\alpha \beta} X^{\beta} \\
	&=\sum_{\mu,I\in\Omega(x^{\mu})} \Lambda^{\mu} X^{\mu}\\
	&=\sum_{\mu,I\in\Omega(x^{\mu})}(1-\lambda^{\mu I})x^{\mu}+\lambda^{\mu I}x^{I}
\end{split}
\end{equation}
where we redefined the sum over the entries $\alpha$ to a double sum over the entries $\mu$ and their corresponding neighbors. Since each data point $x^{\mu}$ is in $k_{\mu}$ other nearest neighbors of some other data points, we can rearrange the terms in this sum to group together the data points $x^{\mu}$. This gives

\begin{equation}
\begin{split}
\sum_{\mu,I\in\Omega(x^{\mu})}(1-\lambda^{\mu I})x^{\mu}+\lambda^{\mu I}x^{I}	&=\sum_{\mu}\left(K-\sum_{I\in\Omega(x^{\mu})}\lambda^{\mu I}\right)x^{\mu}+\sum_{\nu|x^{\mu}\in\Omega(x^{\nu})}\lambda^{\nu J_{\nu}}x^{\mu} \\ 
	&=\sum_{\mu}\left(K-\sum_{I\in\Omega(x^{\mu})}\lambda^{\mu I}+\sum_{\nu|x^{\mu}\in\Omega(x^{\nu})}\lambda^{\nu J_{\nu}}\right)x^{\mu}
\end{split}
\end{equation}
where the last sum here is over the neighborhoods $\Omega(x^{\nu})$ such that $x^{\mu}\in\Omega(x^{\nu})$, with corresponding Latin index $J_{\nu}$. Dividing both sides of this equation by $1/DK$, we find
\begin{equation}\label{eq:mean}
\bar{Y}=\bar{x}+E
\end{equation}
where and overbar denotes sample mean (or sample variance/covariance below) and $E$ is an ``error term" given by 
\begin{equation}\label{eq:errors}
E=\frac{1}{DK}\sum_{\mu}\left(\sum_{\nu|x^{\mu}\in\Omega(x^{\nu})}\lambda^{\nu J_{\nu}}-\sum_{I\in\Omega(x^{\mu})}\lambda^{\mu I}\right)x^{\mu}.
\end{equation}
Here we can see that the sample mean of the synthetic data can be written as the sample mean of the original data with an induced error which explicitly depends on the choice of $\lambda$'s, and implicitly depends on how many neighborhoods each $x^{\mu}$ belongs to. Now we can take the expected value of both sides with respect to both the distribution of $\lambda$'s and the distribution from which the $x$'s are draw, and using the fact that the expected value of the sample mean is equal to the expected value of the random variable, we find 
\begin{align} \label{eq:answer}
\bb{E}(Y)&=\bb{E}(x)+\frac{1}{DK}\sum_{\mu}\bb{E}\left(\left(\sum_{\nu|x^{\mu}\in\Omega(x^{\nu})}\lambda^{\nu J_{\nu}}-\sum_{I\in\Omega(x^{\mu})}\lambda^{\mu I}\right)x^{\mu} \right) \\ 
&=\bb{E}(x)+\frac{1}{D}\sum_{\mu} \left(\sum_{\nu|x^{\mu}\in\Omega(x^{\nu})}\bb{E}\left(\lambda \right)-\sum_{I\in\Omega(x^{\mu})}\bb{E}\left(\lambda \right)\right) \bb{E}(x) \\ 
&=\bb{E}(x)+\bb{E}\left(\lambda \right) \bb{E}(x) \frac{1}{DK}\sum_{\mu} \left(k_{\mu}-K\right) \\
&=\bb{E}(x)
\end{align}
where we have introduced $k_{\mu}$ as the number of nearest neighborhoods to which $x^{\mu}$ belongs. To see why the expected error is zero, consider a directed nearest neighbor graph on the data, where edges point from a data point to each of it's $K$ nearest neighbors.  On this graph, $K$ represents the number of arrows exiting $x^{\mu}$ while $k_{\mu}$ represents the number of arrows entering $x^{\mu}$. When summed over all nodes, these two quantities are equivalent, as they are both different ways of enumerating the edges of the graph. 

This result differs from the analogous result in \cite{smote}. There the authors found
\begin{equation}
\bb{E}(Y)=(1-\bb{E}(\lambda))\bb{E}(x)+\bb{E}(\lambda)\bb{E}(x^{nn}).
\end{equation}
There they argue that for ``symmetric" distributions, $\bb{E}(x^{nn})=\bb{E}(x)$, yielding eq.  \eqref{eq:answer}. Here we show that this result holds regardless of the distribution, so long as the data points are i.i.d. This is a reasonable assumption for many data sets, such as those studied in the main text.

\begin{comment}
 Comparing this to the equivalent calculation in \cite{smote}, we find the surprisingly simple result
\begin{equation}
\bb{E}(y_{nn})= \bb{E}(x) \frac{\bar{k}}{K}.
\end{equation}
This says that the expected value of the nearest neighbors of a random variable $x$ is given by the expected value of $x$ times the ratio of the average number of neighborhoods a given $x^{\mu}$ belongs to the number of neighbors in it's neighborhood. Intuitively this result makes sense, since each point in the data set is drawn from the same distribution. However, data points in dense regions will be over represented in the distribution of nearest neighbors since they may belong to many neighborhoods, while sparse regions will be under represented. Therefore, the expected value over the distribution of nearest neighbors will depend on how many neighborhoods the data points belong to, on average.  The ``symmetric distributions" of \cite{smote} occurs when each data point belongs to $K$ different neighborhoods, on average. 
\end{comment}
\subsection*{Covariance}
The calculation of covariance is more involved, so we will leave out most of the algebra.  For this we let lower case Latin subscripts such as $i$ and $k$ denote individual features of a sample.  The sample covariance can be written as 
\begin{equation}
\overline{\cov}\left(Y_{i}Y_{k}\right)=\frac{1}{DK} \sum_{\alpha}Y_i^{\alpha}Y^{\alpha}_k - \bar{Y_i}\bar{Y_k}.
\end{equation}
In eq. \eqref{eq:mean} we showed that 
\begin{equation}
\bar{Y_i}=\bar{x_i}+E_i
\end{equation}
where $E_i$ is the relevant component of the error term eq.  \eqref{eq:errors}. Plugging in eq. \eqref{compForm} and expanding, we find 
\begin{equation}
\overline{\cov}\left(Y_{i}Y_{k}\right)=\overline{\cov}\left({x_ix_j}\right)+\Sigma
\end{equation}
where 
\begin{equation}\label{coverror}
\begin{split}
\Sigma &=\frac{1}{DK}\sum_{\mu}\left(\sum_{\nu|x^{\mu}\in\Omega(x^{\nu})}\lambda_{i}^{\nu J_{\nu}}\lambda_{k}^{\nu J_{\nu}}+\sum_{I\in\Omega(x^{\mu})}\lambda_{i}^{\mu I}\lambda_{k}^{\mu I}-\sum_{I\in\Omega(x^{\mu})}\left(\lambda_{i}^{\mu I}+\lambda_{k}^{\mu I}\right)\right)x_{i}^{\mu}x_{k}^{\mu} \\
&+\frac{1}{DK}\sum_{\mu} \sum_{I\in\Omega(x^{\mu})}\left(x_{i}^{I}x_{k}^{\mu}\left(\lambda_{i}^{\mu I}-\lambda_{k}^{\mu I}\lambda_{i}^{\mu I}\right)+x_{k}^{I}x_{i}^{\mu}\left(\lambda_{k}^{\mu I}-\lambda_{i}^{\mu I}\lambda_{k}^{\mu I}\right)\right) \\
&+\bar{x}_{i}E_{k}+\bar{x}_{k}E_{i}-E_{i}E_{k}
\end{split}
\end{equation}
is the ``error term".  This equation, while far more opaque than it's counter part in eq. \eqref{eq:errors}, but things simplify when we take the expected value.  Let's start with the last line in the equation above: 

\begin{equation}
\begin{split}
\bb E(\bar{x}_{i}E_{k})	&=\bb E\left[\left(\frac{1}{D}\sum_{\mu}x_{i}\right)\left(\frac{1}{D}\sum_{\mu}\left(\frac{1}{K}\sum_{\nu|x^{\mu}\in\Omega(\x^{\nu})}\lambda_{k}^{\nu J_{\nu}}-\frac{1}{K}\sum_{I\in\Omega(\x^{\mu})}\lambda_{k}^{\mu I}\right)x_{k}^{\mu}\right)\right] \\ 
	&=\bb E\left[\left(\frac{1}{D}\sum_{\mu}x_{i}\right)\left(\frac{1}{D}\sum_{\mu}\left(\frac{k_{\mu}}{K}\bb E(\lambda_{k})-\bb E(\lambda_{k})\right)x_{k}^{\mu}\right)\right]\\
	&=\bb E(\lambda_{k})\bb E\left[\left(\frac{1}{D}\sum_{\mu}x_{i}\right)\left(\frac{1}{D}\sum_{\mu}\frac{k_{\mu}x_{k}^{\mu}}{K}-\frac{1}{D}\sum_{\mu}x_{k}^{\mu}\right)\right]\\
	&=\bb E(\lambda_{k})\left(\bb E\left[\left(\frac{1}{D}\sum_{\mu}x_{i}\right)\left(\frac{1}{D}\sum_{\mu}\frac{k_{\mu}x_{k}^{\mu}}{K}\right)\right]-\mathbb{E}\left[\left(\frac{1}{D}\sum_{\mu}x_{i}\right)\left(\frac{1}{D}\sum_{\mu}x_{k}^{\mu}\right)\right]\right) \\ 
	&=\bb E(\lambda_{k})\bb E\left[\left(\frac{1}{D}\sum_{\mu}x_{i}^{\mu}\right)\left(\frac{1}{D}\sum_{\mu}\frac{k_{\mu}x_{k}^{\mu}}{K}\right)\right]-\bb E(\lambda_{k})\left(\text{cov}(\bar{x}_{i},\bar{x}_{j})+\bb E(\bar{x}_{i})\bb E(\bar{x}_{k})\right)\\
	&=\bb E(\lambda_{k})\bb{E}\left(\bar{x}_i\bar{x}^{nn}_k\right)-\bb E(\lambda_{k})\left(\frac{1}{D}\text{cov}(x_{i},x_{k})+\bb E(x_{i})\bb E(x_{k})\right) 
\end{split}
\end{equation}
where we used the fact that
\begin{align*}
\cov(\bar{x_i},\bar{x_k}) &= \frac{1}{D^2}\sum_{\mu}\sum_{\nu} \cov(x_i^{\mu}, x_k^{\nu} )\\
&= \frac{1}{D^2}\sum_{\mu}\cov(x_i^{\mu}, x_k^{\mu} )\\
&=\frac{1}{D}\cov(x_i x_k)
\end{align*}
when the samples $x^{\mu}$ and $x^{\nu}$ are independent. We can apply this same logic to the first term along with eq. \eqref{eq:answer} to obtain 
\begin{equation}
\bb{E}(\bar{x}_i,\bar{x}^{nn}_k)=\frac{1}{D}\cov(x_i,x_j)+\bb{E}(x_i)\bb{E}(x_k)
\end{equation}
which yields
\begin{align}\label{eq:zero1}
\bb E(\bar{x}_{i}E_{k})=0.
\end{align}
The calculation of $\bb E(\bar{x}_{i}E_{k})$ and $\bb E(E_{i}E_{k})$ in this same way also yield zero.  The only thing left to do is to calculate the first and second terms in $\Sigma$.  For the first term we have 
\begin{align*}
 \bb E & \left(\frac{1}{DK}\sum_{\mu}\left(\sum_{\nu|x^{\mu}\in\Omega(\x^{\nu})}\lambda_{i}^{\nu J_{\nu}}\lambda_{k}^{\nu J_{\nu}}+\sum_{I\in\Omega(\x^{\mu})}\lambda_{i}^{\mu I}\lambda_{k}^{\mu I}-\sum_{I\in\Omega(\x^{\mu})}\left(\lambda_{i}^{\mu I}+\lambda_{k}^{\mu I}\right)\right)x_{i}^{\mu}x_{k}^{\mu}\right)\\
&=\frac{1}{DK}\sum_{\mu}\left(\sum_{\nu|x^{\mu}\in\Omega(\x^{\nu})}\bb E\left(\lambda_{i}^{\nu J_{\nu}}\lambda_{k}^{\nu J_{\nu}}\right)+\sum_{I\in\Omega(\x^{\mu})}\bb E\left(\lambda_{i}^{\mu I}\lambda_{k}^{\mu I}\right)-\sum_{I\in\Omega(\x^{\mu})}\left(\bb E\left(\lambda_{i}^{\mu I}\right)+\bb E\left(\lambda_{k}^{\mu I}\right)\right)\right)\bb E(x_{i}^{\mu}x_{k}^{\mu}) \\ 
&=\frac{1}{DK}\sum_{\mu}\left(\left(k_{\mu}+K\right)\left(\text{var}(\lambda)+\bb E(\lambda)^{2}\right)-2K\bb E(\lambda)\right)\left(\cov(x_{i}x_{k})+\bb E(x_{i})\bb E(x_{k})\right) \\
&=\left(\cov(x_{i}x_{k})+\bb E(x_{i})\bb E(x_{k})\right)\frac{1}{DK}\sum_{\mu}\left(\left(k_{\mu}+K\right)\left(\text{var}(\lambda)+\bb E(\lambda)^{2}\right)-2K\bb E(\lambda)\right) \\ 
&=\left(\cov(x_{i}x_{k})+\bb E(x_{i})\bb E(x+k)\right)\left(\left(\text{var}(\lambda)+\bb E(\lambda)^{2}\right)\frac{1}{DK}\sum_{\mu}(k_{\mu}+K)-2\bb E(\lambda)\right)\\
&=2\left(\cov(x_{i}x_{k})+\bb E(x_{i})\bb E(x_{k})\right)\left(\text{var}(\lambda)+\bb E(\lambda)^{2}-\bb E(\lambda)\right)
\end{align*}
while the second term yields
\begin{align*}
\bb E\left(\frac{1}{DK}\sum_{\mu}\sum_{I\in\Omega(\x^{\mu})}x_{i}^{I}x_{k}^{\mu}\left(\lambda_{i}^{\mu I}-\lambda_{k}^{\mu I}\lambda_{i}^{\mu I}\right)\right)+ (i\leftrightarrow k)	&=\frac{1}{DK}\sum_{\mu}\sum_{I\in\Omega(x^{\mu})}\bb E(x_{i}^{I}x_{k}^{\mu})\left(\bb E(\lambda)-\bb E(\lambda^{2})\right) + (i\leftrightarrow k)\\
	&=\frac{1}{DK}\sum_{\mu}\sum_{I\in\Omega(x_{\mu})}\bb E(x_{i})\bb E(x_{k})\left(\bb E(\lambda)-\bb E(\lambda^{2})\right)+ (i\leftrightarrow k)\\
	&=2\bb E(x_{i})\bb E(x_{k})\left(\bb E(\lambda)-\text{var}(\lambda)-\bb E(\lambda)^{2}\right)
\end{align*}
where we used the fact that $x^{\mu}$ and it's nearest neighbors are independent since they are independent samples. All together, we have
\begin{equation}
\bb{E}(\Sigma)=2\cov(x_{i}x_{k})\left(\text{var}(\lambda)+\bb E(\lambda)^{2}-\bb E(\lambda)\right)
\end{equation}
which, along with eq. \eqref{eq:answer} gives the final result: 
\begin{equation}\label{eq:finCov}
\cov \left( Y_i,Y_k\right)=2\cov(x_i,x_k)\left(\text{var}(\lambda)+\bb E(\lambda)^{2}-\bb E(\lambda)\right).
\end{equation}
If $\lambda$ is drawn from $I_{[0,1]}$, then we obtain the result
\begin{equation}\label{eq:finCovUni}
\cov \left( Y_i,Y_k\right)=\frac{2}{3}\cov(x_i,x_k)
\end{equation}
which agrees with the result of \cite{smote} when the distribution is symmetric and/or high dimensional. Here, as in eq. \eqref{eq:answer}, we show that this result holds when the data points are i.i.d regardless of dimensionality and symmetry properties.

\bibliography{tsmote}
\bibliographystyle{plain}

\end{document}